\documentclass[10pt,twocolumn,letterpaper]{article}

\usepackage{cvpr}              % To produce the CAMERA-READY version
\usepackage{bm}
\usepackage[accsupp]{axessibility}  % Improves PDF readability for those with disabilities.

\definecolor{mygreen}{rgb}{0,0.7,0}

\definecolor{cvprblue}{rgb}{0.21,0.49,0.74}
\usepackage[pagebackref,breaklinks,colorlinks,allcolors=cvprblue]{hyperref}
\definecolor{myyellow}{rgb}{0.8,0.6,0.0}

\usepackage{multirow}

\def\paperID{5746} % *** Enter the Paper ID here
\def\confName{CVPR}
\def\confYear{2026}

\title{Group-DINOmics: Incorporating People Dynamics into DINO \\ for Self-supervised Group Activity Feature Learning}
\if 0
\author{Ryuki Tezuka\\
Institution1\\
Institution1 address\\
{\tt\small firstauthor@i1.org}
\and
Second Author\\
Institution2\\
First line of institution2 address\\
{\tt\small secondauthor@i2.org}
\and
Second Author\\
Institution3\\
First line of institution3 address\\
{\tt\small secondauthor@i3.org}
}
\fi

\author{Ryuki Tezuka ~~ Chihiro Nakatani ~~ Norimichi Ukita\\
Toyota Technological Institute\\
{\tt\small \{sd24437,sd23501,ukita\}@toyota-ti.ac.jp}
}

\if 0
\author{Ryuki Tezuka ~~ Chihiro Nakatani ~~ Norimichi Ukita\\
Toyota Technological Institute\\
{\tt\small rkrk.1536.tktk@gmail.com ~~ \{sd23501,ukita\}@toyota-ti.ac.jp}
}
\fi

\begin{document}
\maketitle
\begin{abstract}
This paper proposes Group Activity Feature (GAF) learning without group activity annotations. 
Unlike prior work, which uses low-level static local features to learn GAFs, we propose leveraging dynamics-aware and group-aware pretext tasks, along with local and global features provided by DINO, for group-dynamics-aware GAF learning.
To adapt DINO and GAF learning to local dynamics and global group features, our pretext tasks use person flow estimation and group-relevant object location estimation, respectively.
Person flow estimation is used to represent the local motion of each person, which is an important cue for understanding group activities. 
In contrast, group-relevant object location estimation encourages GAFs to learn scene context (e.g., spatial relations of people and objects) as global features.
Comprehensive experiments on public datasets demonstrate the state-of-the-art performance of our method in group activity retrieval and recognition. Our ablation studies verify the effectiveness of each component in our method.
Code:~\url{https://github.com/tezuka0001/Group-DINOmics}. 

\end{abstract}
\section{Introduction}
\label{sec:intro}

Group activity analysis is an important topic in various fields, such as sports analytics~\cite{zhao2025survey, DBLP:series/sbcs/Fujii25}, robotics~\cite{DBLP:conf/cvpr/JRDB-Social}, and anomaly detection in surveillance videos~\cite{DBLP:conf/cvpr/Surveillance}.

Supervised group activity recognition has been widely investigated~\cite{DBLP:conf/iccv/SBGAR, DBLP:conf/cvpr/CRM, DBLP:conf/cvpr/Detector_Free, DBLP:conf/cvpr/Dual-AI, DBLP:conf/eccv/flow_assist, DBLP:conf/eccv/NBA, DBLP:conf/eccv/skeleton-based, DBLP:conf/iccv/DIN, DBLP:conf/wacv/LiGAR, DBLP:conf/cvpr/ChappaNNSLDL23, DBLP:journals/access/SOGAR, DBLP:journals/mva/POGARS, DBLP:journals/pami/HiGCIN, DBLP:journals/tcsv/KRGFormer, DBLP:conf/mva/GAR_grouping, DBLP:conf/iccv/GroupFormer, DBLP:conf/eccv/Composer, DBLP:conf/eccv/SACRF, DBLP:conf/cvpr/AT, DBLP:conf/cvpr/ARG, DBLP:conf/cvpr/SSU}.
These methods are trained with annotations for group activity classes to classify each video into one of the predefined classes.
However, supervised group activity recognition has two difficulties: (i) the need for a large number of manual annotations and (ii) the need to predefine complex and possibly unknown group activity classes.
%(i) a large number of manual annotations are needed, and (ii) it assumes that complex and potentially unknown group activity classes are predefined.

To avoid the two difficulties, HRN~\cite{DBLP:conf/eccv/HRN} and GAFL~\cite{DBLP:conf/cvpr/GAFL} learn Group Activity Features (GAFs) as compact latent vectors in a self-supervised manner.
In~\cite{DBLP:conf/eccv/HRN,DBLP:conf/cvpr/GAFL}, pretext tasks that require neither manual annotations nor predefined group activity classes are designed to learn GAFs.
The GAFs can be used for group activity retrieval and pretraining for group activity recognition, as demonstrated in~\cite{DBLP:conf/cvpr/GAFL}.

\begin{figure}[t]
    \centering
    \includegraphics[width=\columnwidth]{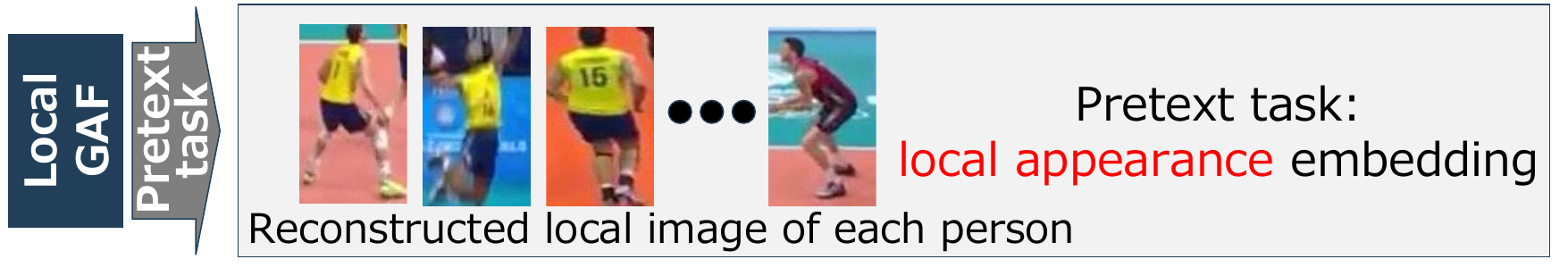}\\
    \vspace*{-1mm}
    (1) Previous methods~\cite{DBLP:conf/eccv/HRN, DBLP:conf/cvpr/GAFL}\\
    \vspace*{2mm}
    \includegraphics[width=\columnwidth]{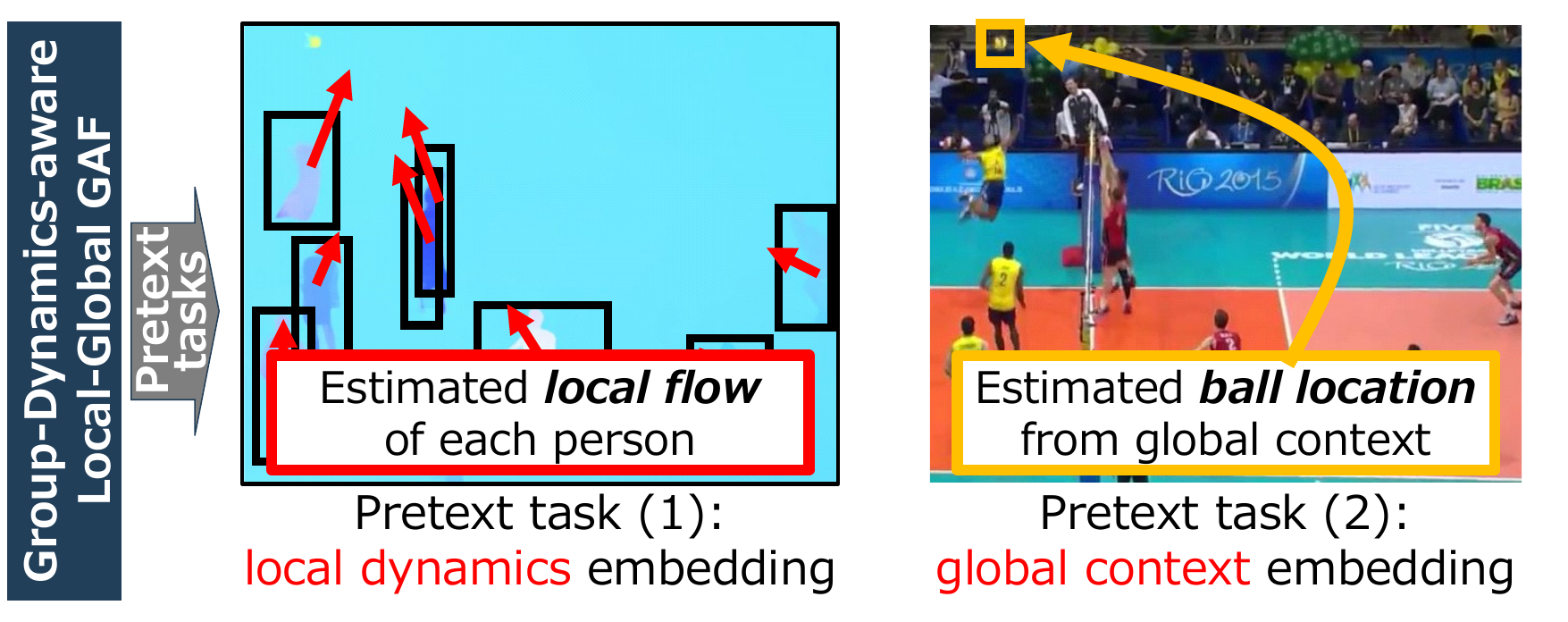}\\
    \vspace*{-1mm}
    (2) Ours for local dynamics and global context learning\\
    \vspace*{-1mm}
    \caption{Our self-supervised GAF learning augmented by two pretext tasks: (1) person flow estimation for local dynamics embedding into GAFs and (2) group-relevant object localization for global context embedding into GAFs. Compared with previous self-supervised methods~\cite{DBLP:conf/eccv/HRN, DBLP:conf/cvpr/GAFL} that utilize only local appearance features, our pretext tasks enhance GAF learning.}
    \label{fig:teaser}
    \vspace*{-2mm}
\end{figure}

However, the GAFs extracted by these methods~\cite{DBLP:conf/eccv/HRN, DBLP:conf/cvpr/GAFL} are not sufficiently group-activity-aware for two reasons. (1) Only local appearance cues are embedded into GAFs because these GAFs are trained via local image reconstruction as a pretext task. In contrast, the motion features of each person, which capture local dynamics related to activities, are more beneficial,
% ..... can be removed "for group activity understanding,"
as demonstrated in supervised group activity recognition~\cite{DBLP:conf/cvpr/AT,DBLP:conf/iccv/GroupFormer}. (2) Only the local features of each person are embedded into GAFs, whereas global activity-aware features that capture the spatial configuration among people are important~\cite{DBLP:conf/cvpr/CRM,DBLP:conf/cvpr/SSU}.
% .... can be remove: ", even in the form of a compact latent vector,"
%To encourage self-supervised GAFs, even in the form of a compact latent vector, to maintain a good balance between these local dynamics and global context features, this paper focuses on image feature extractors and pretext tasks.
To ensure that GAFs learned in a self-supervised manner, even when represented as compact latent vectors, maintain a good balance between local dynamics and global context features, we focus on image feature extractors and pretext tasks.

{\bf Image feature extractors.}
Even large-scale pretrained feature extractors often exhibit a bias toward either local or global features.
For example,
%like convolution-based extractors~\cite{DBLP:conf/iclr/VGG}, 
pixelwise reconstruction-based feature extractors~\cite{DBLP:conf/cvpr/MAE} are biased toward local features~\cite{DBLP:conf/iclr/hybrid_disti}.
CLIP~\cite{DBLP:conf/icml/CLIP} learns global features through image-text alignment, but its representation of local features remains insufficient~\cite{DBLP:conf/eccv/clip_dinoiser}.
On the other hand, the DINO family~\cite{DBLP:conf/iccv/DINO, DBLP:journals/tmlr/DINOv2, DINOv3} jointly learns local and global features.
%by aligning local and global views. 
In particular,
%through several methodological refinements (e.g., a patchwise local representation trained by iBOT~\cite{DBLP:conf/iclr/Zhou0W0XYK22} and its augmentation using gram anchoring), 
%DINOv3~\cite{DINOv3} achieves a balance between local and global features.
DINOv3~\cite{DINOv3} balances local and global features.
%Since both local and global features are essential for group activity representation, as mentioned above, 
%%%DINOv3 is, therefore, a promising choice for GAF learning.

{\bf Pretext tasks.}
However, even DINOv3, when applied as is, is insufficient for GAF learning because it is trained only on still images for general-purpose features.
Such general-purpose features include many local features (e.g., clothing textures) and global features (e.g., floor textures), neither of which is relevant to group activities.
In addition, temporal features, such as motion dynamics, are not explicitly learned from still images alone.
To adapt DINOv3 to group activity analysis with local dynamics and global context features in a self-supervised manner, we propose two pretext tasks that take a GAF as input (Fig.~\ref{fig:teaser}).
(1) Person flow estimation encourages GAFs to include the local dynamics features of each person.
(2) Group-relevant objects (e.g., a ball in team sports) are localized 
%without their local image cues
to encourage GAFs to include scene context (e.g., spatial relations of people, foreground, and background objects) as global features.

Our contributions are summarized as follows:
\begin{itemize}
    \item {\bf Pretext tasks enhancing local-global GAFs:}
    \begin{itemize}
        \item
        {\bf Person flow estimation:}
        The local dynamics features of each person are embedded into GAFs via person flow estimation. %However, unlike appearance features~\cite{DBLP:conf/eccv/HRN,DBLP:conf/cvpr/GAFL}, dynamics feature embedding is difficult with DINOv3, as DINOv3 is trained on still images. We complement flow estimation from GAFs with an auxiliary flow estimation branch, enabling DINOv3 to learn motion dynamics directly.
        %In addition to flow estimation from a GAF, therefore, auxiliary flow estimation is used to allow DINOv3 to directly learn dynamics.
        To address a difficulty in adapting DINOv3 to dynamics features, our method complements flow estimation from GAFs with an auxiliary flow estimation branch, enabling DINOv3 to learn motion dynamics directly.
        
        \item
        {\bf Group-relevant object localization with inpainting:}
        %Group-relevant objects are localized from a GAF.
        %To encourage GAFs to learn the spatial relations of people and group-relevant objects as global features, these objects are localized from a GAF.
        Group-relevant objects are localized from a GAF to learn the global spatial relations between people and objects.
        This paper proposes inpainting these objects during training, so that our method cannot rely on their local appearance cues and must estimate their locations from global features included in a GAF.
    \end{itemize}

    %These pretext tasks can be self-supervised with pseudo person flows and object locations estimated by a flow estimator and object detectors, respectively.
    These pretext tasks can be trained in a self-supervised manner using pseudo labels for person flows and object locations estimated by a flow estimator and object detectors, respectively.
    %because person flows and object locations are obtained by a flow estimator and object detectors, respectively.
    Thus, as with self-supervised GAF learning~\cite{DBLP:journals/pr/WangLLZGFC25,DBLP:journals/pr/WangLLGWGF24}, which also uses such estimated inputs (i.e., human poses in~\cite{DBLP:journals/pr/WangLLZGFC25,DBLP:journals/pr/WangLLGWGF24}), our pretext tasks maintain the self-supervised nature, which requires no manual annotations of group activities.
    %%% .... While self-supervised learning with these pseudo labels, including our method and~\cite{DBLP:journals/pr/WangLLZGFC25,DBLP:journals/pr/WangLLGWGF24}, requires additional pretrained models to obtain these pseudo labels, it is superior to self-supervised learning with no pseudo labels~\cite{DBLP:conf/cvpr/ChappaNNSLDL23,DBLP:journals/access/SOGAR} even if the pseudo labels are obtained by general-purpose models, as validated in our experiments.

    \item {\bf Comprehensive analysis of GAFs trained with our method:} Our experiments show that the combination of DINOv3 and our pretext tasks outperforms other methods in the group activity retrieval and recognition tasks.
\end{itemize}
\begin{figure*}[t]
    \centering
     \includegraphics[width=\textwidth]{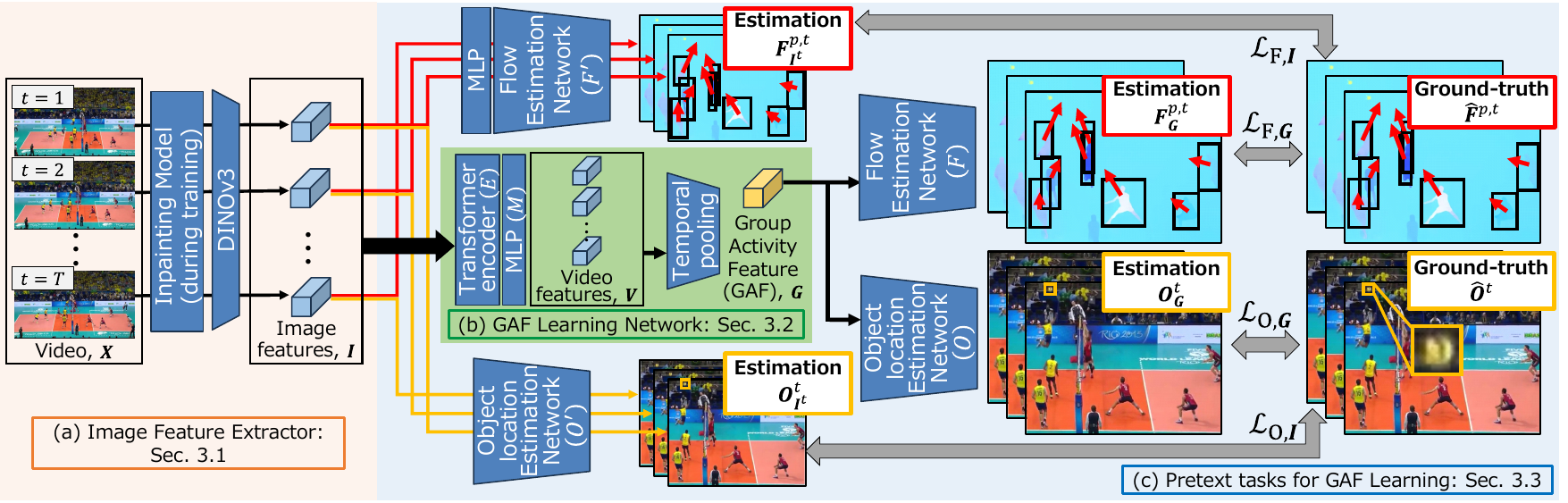}
     \vspace*{-6mm}
     \caption{Overview of our network. (a) Image feature extractor. Group-relevant objects
     %(i.e., ball)
     are inpainted to enhance global feature learning in (c).
     (b) GAF learning network. Image features are fed into the transformer encoder, MLP, and temporal pooling to obtain a GAF. (c) Pretext tasks for GAF learning. The flow of each person and the locations of the group-relevant objects are estimated from the GAF.}
     \label{fig:overview_network}
\end{figure*}

\section{Related Work}
\label{sec:rel_work}

\subsection{Supervised Group Activity Recognition}
\label{subsec:sgar}

Local features, such as the behavior of each person, are important to capture group activities. To learn such local features, 
%optical flows~\cite{DBLP:conf/cvpr/AT, DBLP:conf/iccv/GroupFormer, DBLP:conf/eccv/flow_assist} and 
the location of each person~\cite{DBLP:conf/eccv/huntingGAR} and 
individual action labels~\cite{DBLP:conf/cvpr/ARG, DBLP:conf/eccv/huntingGAR} are estimated as auxiliary tasks within supervised methods.
%These results suggest that recognizing each person's local features helps group activity understanding.
%
In addition to these local features, global features are also informative for people's interactions. Previous supervised methods learn such interactions by estimating the spatial configuration of people~\cite{DBLP:conf/cvpr/CRM, DBLP:conf/cvpr/SSU}.
%%% (object inputs) or group-relevant objects (e.g., a ball and a net in sports)~\cite{DBLP:conf/eccv/skeleton-based, DBLP:journals/mva/POGARS, DBLP:journals/pr/PerezLK22}.
%In all of these supervised methods, complex relations are learned through group activity annotations as strong supervision.
%%%% ..... WSGAR 
%While the methods mentioned above use additional supervision in addition to group activity labels, weakly-supervised methods work only with per-video group activity labels~\cite{DBLP:conf/cvpr/Detector_Free,DBLP:conf/eccv/NugrohoWLPWKK24,DBLP:conf/eccv/NBA,DBLP:journals/corr/abs-2502-09967}.

Unlike these supervised recognition methods, our method learns GAFs, instead of recognizing group activity classes, without any manual annotations of group activity classes in a self-supervised manner described below.
%
%%% .... (deleted) Since such self-supervised learning is more challenging than supervised learning, the auxiliary tasks employed in these supervised methods are insufficient.

\subsection{Self-supervised Feature Learning}
\label{subsec:sfeat}

\noindent{\bf General image feature learning.}
In many pretext tasks for self-supervised feature learning~\cite{DBLP:conf/cvpr/Context_Encoders,DBLP:conf/cvpr/MAE,DBLP:conf/cvpr/SimMIM}, images are reconstructed from the features of their masked images to learn general features.
%%%% .... (can be removed) so that the features are trained to be essential latent features.
% 
Unlike these general methods, our method is designed to extract features optimized for 
%understanding 
group activities through group-activity-aware pretext tasks.

\noindent{\bf Group activity feature learning.}
HRN~\cite{DBLP:conf/eccv/HRN} and GAFL~\cite{DBLP:conf/cvpr/GAFL} train GAFs through a pretext task that reconstructs the local appearance features (e.g., color and texture) of each person from a GAF.
On the other hand, our method learns activity-aware features (i.e., person flows and group-relevant object locations) via our proposed pretext tasks.

\noindent{\bf Dynamics feature learning.}
%with flows.}
Optical flows can serve as dynamics features that are activity-aware.
Flow estimation is used as a pretext task for self-supervised pre-training of image feature extractors in individual action recognition~\cite{DBLP:conf/eccv/DjVu, DBLP:conf/iccv/dense_flow, DBLP:conf/cvpr/Im2Flow, DBLP:conf/cvpr/self_flow}.
These methods estimate dense pixelwise flows across the entire image to represent detailed full-body motions.
%of one or a small number of people. 
%However, such dense flows are not suitable for independently representing the motions of widely dispersed people in group activities without being interfered with by meaningless background flows. Therefore, our pretext task estimates local per-person flows from a GAF.
On the other hand, our pretext task estimates local per-person flows from a GAF to focus on the activity-aware features of each person.
%without being interfered with by meaningless background flows.

Our method uses DINOv3 for feature extraction. Among temporally-aware DINO optimization methods~\cite{DBLP:journals/corr/abs-2412-11673,DBLP:conf/iccv/SalehiGSA23,DBLP:conf/cvpr/KimCHYLK25,DBLP:conf/eccv/DingQXLX24}, FlowDINO~\cite{FlowDINO-iros2023} and DINO-tracker~\cite{DBLP:conf/eccv/TumanyanSBD24} finetune DINO with optical flows in a self-supervised manner. While they use the flows to learn temporal consistency between the DINO features of sequential frames, our pretext task encourages DINO to learn flows as dynamics features directly.

%%% previous: spatial features, ours non-spatial features (for compact features. why? because of a limited number of training data)

\noindent{\bf People-object interaction learning.}
Since people interact with objects in many group activities (e.g., a ball and a net in sports), such people-object interactions can be global cues in group activity recognition~\cite{DBLP:conf/eccv/skeleton-based, DBLP:journals/mva/POGARS, DBLP:journals/pr/PerezLK22}. Whereas the tracklets of people and objects are simply fed into
%a feature extractor in
these supervised methods~\cite{DBLP:conf/eccv/skeleton-based, DBLP:journals/mva/POGARS, DBLP:journals/pr/PerezLK22}, our pretext task estimates people-object interactions to embed global-activity-aware features into GAFs in a self-supervised manner.
%As also mentioned in Sec.~\ref{subsec:sgar}, since self-supervised learning is more difficult than supervised learning, this pretext task should be sufficient to embed the global features into GAFs.
\section{Proposed Method}
\label{sec:method}

Figure~\ref{fig:overview_network} shows the overview of our network. It extracts a $D$-dimensional GAF, $\bm{G} \in\mathbb{R}^{D}$, from a video.
For this GAF extraction, each frame in the video is fed into an image feature extractor (i.e., DINOv3) independently to extract the features of each frame (Fig.~\ref{fig:overview_network} (a) and Sec.~\ref{subsec:img_feat_ext}).
The set of these image features, $\bm{I}$, is fed into a transformer encoder followed by an MLP to obtain video features. The set of these video features, $\bm{V}$, is pooled along a temporal axis to obtain $\bm{G}$ (Fig.~\ref{fig:overview_network} (b) and Sec.~\ref{subsec:GAF_learning}). During training, for our pretext tasks, $\bm{G}$ is used to estimate the $xy$ flow values of each person and the $xy$ coordinates of group-relevant objects to encourage $\bm{G}$ to learn group activities (Fig.~\ref{fig:overview_network} (c) and Sec.~\ref{subsec:losses}).
Note that we assume that ``the person flows of all people'' and ``the bounding boxes of all people and group-relevant objects'' are given by a flow estimator and generic object detectors (e.g., RAFT~\cite{DBLP:conf/eccv/raft} and YOLOX~\cite{DBLP:journals/corr/YOLOX}), respectively, only during training, not in inference.
%as mentioned in Sec.~\ref{sec:intro}.

%%%%%%%%%%%%%%%%%%%%%%%%%%%%%%%%%%%%%%%%%%%%%%%%%%

\subsection{Image Feature Extractor}
\label{subsec:img_feat_ext}

\subsubsection{Inpainting}
\label{subsubsec:inpainting}

\begin{figure}[t]
    \centering
     \includegraphics[width=\columnwidth]{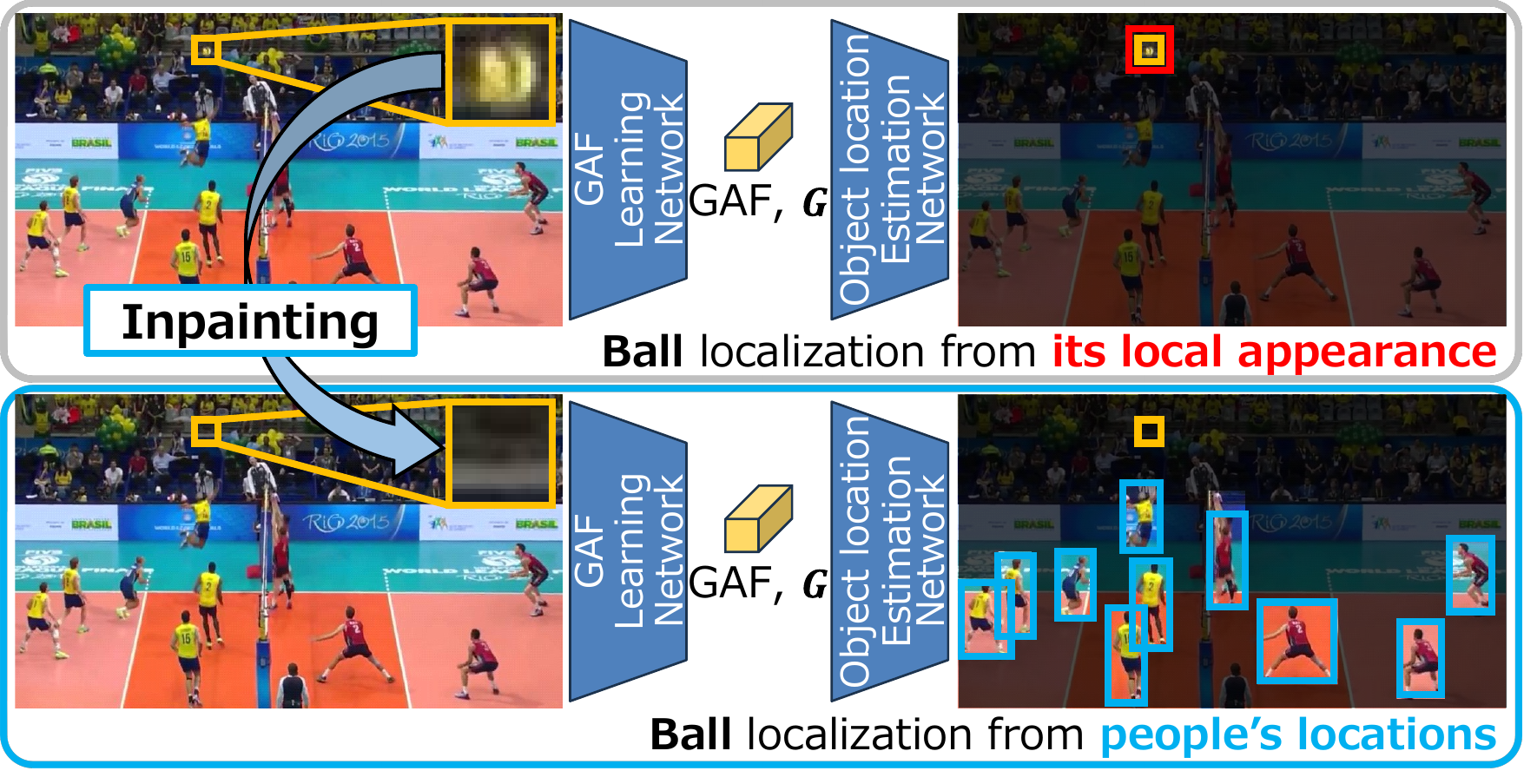}
     \vspace*{-7mm}
     \caption{Inpainting to enhance global feature embedding into a GAF by localizing group-relevant objects (i.e., ball in our method).
     %from the global context represented by people. 
     Top: w/o inpainting. Bottom: w/ inpainting (our method).}
     \label{fig:overview_inpainting}
\end{figure}

During training, each frame of size $W \times H$ pixels in an input video with $T$ frames, $\bm{X} \in\mathbb{R}^{T \times H \times W \times 3}$, is fed into an inpainting model to remove group-relevant objects. 
With $\bm{G}$ extracted from these inpainted images, our pretext task for global feature embedding estimates the locations of these objects at each frame.
% 
%As shown in Fig.~\ref{fig:overview_inpainting}, 
The motivation of this inpainting is to encourage $\bm{G}$ to localize group-relevant objects not from their local appearance cues (i.e., the top of Fig.~\ref{fig:overview_inpainting}) but from global context, including the spatial configuration of people (i.e., the bottom of Fig.~\ref{fig:overview_inpainting}). This allows $\bm{G}$ to learn the global features of a video for representing group activities.
To prevent the inpainting model from learning the appearance of the group-relevant objects, the inpainting model is frozen.

\begin{figure}[t]
% \begin{figure*}[t]
    \centering
     \includegraphics[width=\columnwidth]{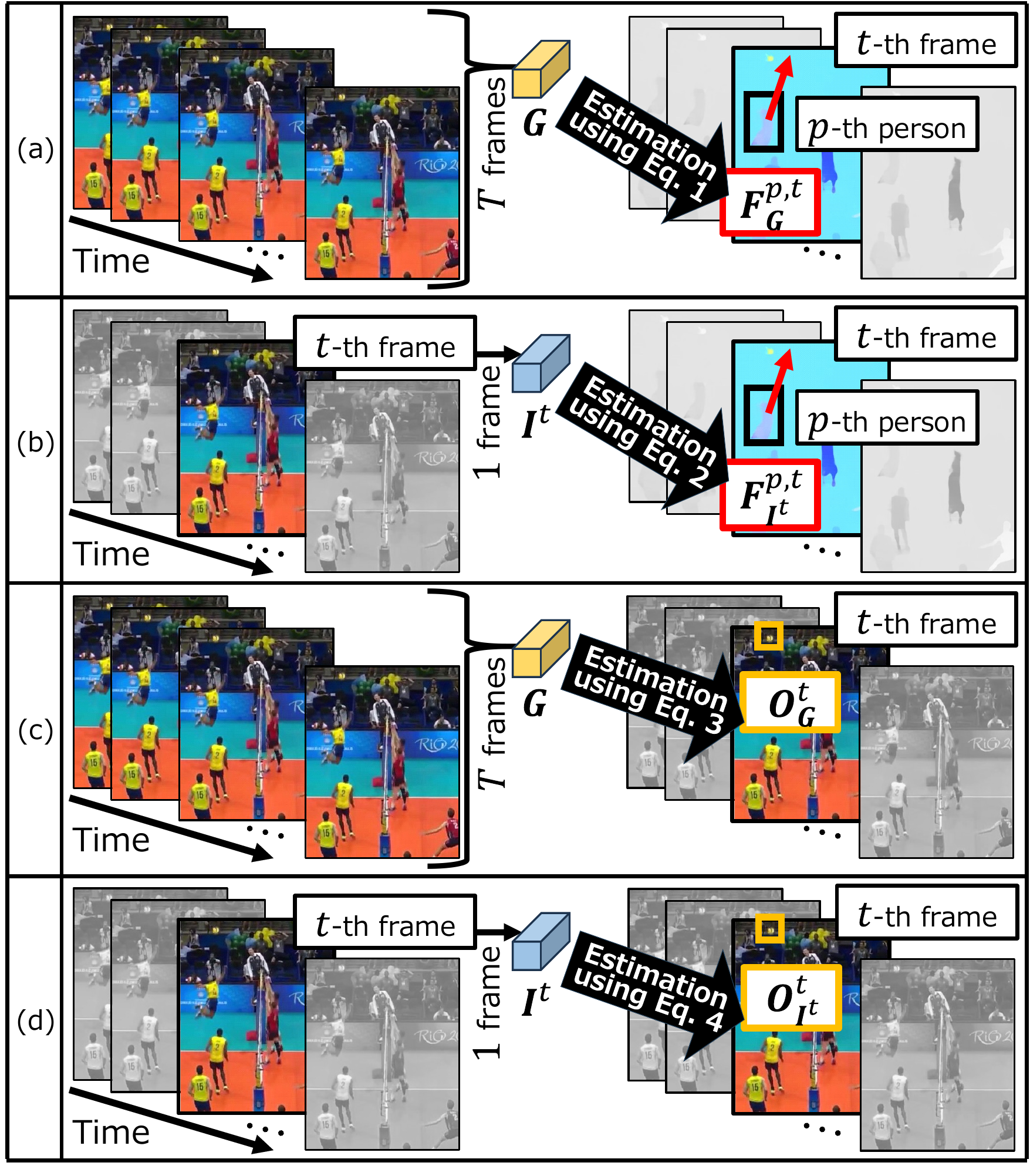}
     \vspace{-7mm}
     \caption{Pretext tasks: overview of person flow estimation, (a) and (b), and group-relevant object location estimation, (c) and (d).
     $\bm{L}^{p,t}$ and $\bm{T}^{t}$ are omitted for simplicity.
     }
     \label{fig:overview_4_losses}
     \vspace*{-1mm}
\end{figure}
% \end{figure*}

\subsubsection{Feature Extraction}
\label{subsec:feat_ext}

Each frame and its inpainted image are fed into DINOv3 to extract image features, $\bm{I} \in\mathbb{R}^{T \times D}$, in inference and training, respectively.
While the previous self-supervised GAF learning methods~\cite{DBLP:conf/eccv/HRN, DBLP:conf/cvpr/GAFL} take person bounding boxes cropped from an image, DINOv3 takes the whole image to capture the global context in our method, as with recent group activity recognition~\cite{DBLP:conf/cvpr/Detector_Free, DBLP:conf/eccv/flow_assist}; our method uses the person bounding boxes only in loss functions.
Following \cite{DBLP:conf/icml/Explora}, our method
%efficiently
finetunes DINOv3 by training only the last two blocks of its ViT.

%%%%%%%%%%%%%%%%%%%%%%%%%%%%%%%%%%%%%%%%%%%%%%%%%%

\subsection{Group Activity Feature Learning Network}
\label{subsec:GAF_learning}

To obtain video features, $\bm{V} \in\mathbb{R}^{T \times D}$, $\bm{I}$ is fed into a transformer encoder~\cite{DBLP:conf/nips/transformer}, $E$, and MLP layers, $M$:
$\bm{V} = M(E(\bm{I}))$.
%\begin{align}
%    \bm{V} = M(E(\bm{I}))
%    \label{eq:trans}
%\end{align}
%
$\bm{V}$ is pooled over time to obtain $\bm{G}$, as in~\cite{DBLP:conf/cvpr/GAFL}.

%In inference, $\bm{G}$ can be used for various tasks, such as group activity retrieval and recognition.
%
%During training, 
``$E$ and $M$'' and DINOv3 are trained from scratch and finetuned, respectively, to learn GAFs through our two pretext tasks (i.e., person flow estimation and group-relevant object location estimation), as described in Sec.~\ref{subsec:losses}.

%%%%%%%%%%%%%%%%%%%%%%%%%%%%%%%%%%%%%%%%%%%%%%%%%%

\subsection{Pretext tasks for GAF Learning}
\label{subsec:losses}

%As pretext tasks, this paper proposes the person flow and group-relevant object estimation tasks, which are described in Sec.~\ref{subsubsec:flow_predition} and Sec.~\ref{subsubsec:group_object_estimation}, respectively.
Our proposed pretext tasks, person flow estimation and group-relevant object location estimation, are described in Sec.~\ref{subsubsec:flow_predition} and Sec.~\ref{subsubsec:group_object_estimation}, respectively.
The overview of these pretext tasks is shown in Fig.~\ref{fig:overview_4_losses}. 

%%%%%%%%%%%%%%%%%%%%%%%%%%%%%%

\subsubsection{Person Flow Estimation}
\label{subsubsec:flow_predition}

\noindent {\bf Estimation by $\bm{G}$.} 
Using $\bm{G}$, the $xy$ flow values of $p$-th person at $t$-th frame, $\bm{F}_{\bm{G}}^{p,t} \in\mathbb{R}^{2}$, are estimated from its location encoding features, $\bm{L}^{p,t} \in\mathbb{R}^{D}$, and the temporal encoding features, $\bm{T}^{t} \in\mathbb{R}^{D}$, as follows (Fig.~\ref{fig:overview_4_losses} (a)):
\begin{align}
\bm{F}_{\bm{G}}^{p,t} = F(\bm{G} + \bm{L}^{p,t} + \bm{T}^{t}),
\label{eq:flow_vid}
\end{align}
where $F$ denotes a flow estimation network, which is a multi-layer perceptron. $F$ is trained from scratch with this pretext task.
$\bm{L}^{p,t}$ and $\bm{T}^{t}$ are obtained by sinusoidal positional and temporal encodings, respectively.
%%% .... (can be omitted) $\bm{L}^{p,t}$ takes the $xy$ coordinates of the center of each person's bounding box, as with~\cite{DBLP:conf/cvpr/GAFL}.

\noindent {\bf Estimation by $\bm{I}^{t}$.}
However, it is not easy for the aforementioned pretext task using $\bm{G}$ to optimize DINOv3 for dynamics features. This is because (i) DINOv3 is trained only on still images, (ii) the backpropagation path toward DINOv3 is long, and (iii) joint learning with the other networks (i.e., $F$, $M$, and $E$) is not simple.
To resolve these difficulties, this paper proposes an auxiliary pretext task using person flow estimation.
Unlike the one using $\bm{G}$ (Eq.~\ref{eq:flow_vid}), this auxiliary task estimates the flow
%of each person
from $\bm{I}^{t}$, which DINOv3 provides, to resolve the above difficulties as follows.
(i) While $\bm{G}$ includes temporal features because $\bm{G}$ is obtained from $\bm{I} = \{ \bm{I}^{1}, \cdots, \bm{I}^{T} \}$, flow estimation only from $\bm{I}^{t}$ requires DINOv3 to embed temporal motion features into $\bm{I}^{t}$.
(ii, iii) The backpropagation path and joint learning are shortened and simplified because the large GAF learning network is not used in this auxiliary task.

The flow values of each person (denoted by $\bm{F}^{p,t}_{\bm{I}^{t}}$) are estimated from $\bm{I}^{t}$ and $\bm{L}^{p,t}$ as follows (Fig.~\ref{fig:overview_4_losses} (b)):
\begin{align}
\bm{F}_{\bm{I}^{t}}^{p,t} = F'(\bm{I}^{t} + \bm{L}^{p,t})
\label{eq:flow_img}
\end{align}
While the network architecture of $F'$ is the same as that of $F$ in Eq.~\ref{eq:flow_vid}, they have different weights.

%%%%%%%%%%%%%%%%%%%%%%%%%%%%%%

\subsubsection{Group-relevant Object Location Estimation}
\label{subsubsec:group_object_estimation}
 
%As shown in Fig.~\ref{fig:overview_network} (c), 
For global context embedding, the locations of group-relevant objects at $t$, $\bm{O}_{\bm{G}}^{t} \in\mathbb{R}^{2N_{o}}$, where $N_{o}$ denotes the number of the group-relevant objects, are estimated from $\bm{G}$ and the temporal encoding features as follows (Fig.~\ref{fig:overview_4_losses} (c)):
\begin{align}
\bm{O}_{\bm{G}}^{t} = O(\bm{G} + \bm{T}^{t}),
\label{eq:obj_vid}
\end{align}
where $O$ is an object location estimation network, which is a multi-layer perceptron, as with $F$.

\noindent {\bf Estimation by $\bm{I}^{t}$.}
As with person flow estimation described in Sec.~\ref{subsubsec:flow_predition}, object location estimation using $\bm{G}$ is insufficient to train $\bm{G}$ to include motion-aware global features.
Therefore, the locations of the group-relevant objects (denoted by $\bm{O}^{t}_{\bm{I}^{t}}$) are also estimated from $\bm{I}^{t}$ (Fig.~\ref{fig:overview_4_losses} (d)):
\begin{align}
\bm{O}^{t}_{\bm{I}^{t}} = O'(\bm{I}^{t})
\label{eq:obj_img}
\end{align}
As with $F$ in Eq.~\ref{eq:flow_vid} and $F'$ in Eq.~\ref{eq:flow_img}, $O$ in Eq.~\ref{eq:obj_vid} and $O'$ in Eq.~\ref{eq:obj_img} have different weights but the same architecture.

%%%%%%%%%%%%%%%%%%%%%%%%%%%%%%

\subsubsection{Loss Functions}
\label{subsubsec:loss}

The whole network is optimized by using the flow loss function, $\mathcal{L}_{\mathrm{F}}$, and the group-relevant object loss function, $\mathcal{L}_{\mathrm{O}}$.

\noindent {\bf Flow loss.}
$\mathcal{L}_{\mathrm{F}}$ is defined as $\mathcal{L}_{\mathrm{F}} = \mathcal{L}_{\mathrm{F}, \bm{G}} + \mathcal{L}_{\mathrm{F}, \bm{I}}$.
$\mathcal{L}_{\mathrm{F}, \bm{G}}$ and $\mathcal{L}_{\mathrm{F}, \bm{I}}$ are computed using $\bm{G}$ and $\bm{I}^{t}$, respectively, as follows: 
\begin{align}
    \mathcal{L}_{\mathrm{F}, \bm{G}} = {\textstyle \sum_{p=1}^P \sum_{t=1}^T \mathcal{L}_{\mathrm{MSE}}\bigl(\bm{F}_{\bm{G}}^{p,t}, \hat{\bm{F}}^{p,t}\bigr)}, \label{eq:flow_loss_g} \\
    \mathcal{L}_{\mathrm{F}, \bm{I}} = {\textstyle \sum_{p=1}^P \sum_{t=1}^T
    \mathcal{L}_{\mathrm{MSE}}\bigl(\bm{F}_{\bm{I}^{t}}^{p,t}, \hat{\bm{F}}^{p,t}\bigr)}, \label{eq:flow_loss_i} 
\end{align}
where $\hat{\bm{F}}^{p,t}$ is the ground-truth flow of $p$-th person at $t$.

\noindent {\bf Group-relevant object loss.}
$\mathcal{L}_{\mathrm{O}}$ is defined as $\mathcal{L}_{\mathrm{O}} = \mathcal{L}_{\mathrm{O}, \bm{G}} + \mathcal{L}_{\mathrm{O}, \bm{I}}$.
$\mathcal{L}_{\mathrm{O}, \bm{G}}$ and $\mathcal{L}_{\mathrm{O}, \bm{I}}$ are computed using $\bm{G}$ and $\bm{I}^{t}$, respectively, as follows:
\begin{align}
    \mathcal{L}_{\mathrm{O}, \bm{G}} = {\textstyle \sum_{t=1}^T \mathcal{L}_{\mathrm{MSE}}\bigl(\bm{O}_{\bm{G}}^{t}, \hat{\bm{O}}^{t}\bigr)}, \label{eq:object_loss_g} \\
    \mathcal{L}_{\mathrm{O}, \bm{I}} = {\textstyle \sum_{t=1}^T
    \mathcal{L}_{\mathrm{MSE}}\bigl(\bm{O}_{\bm{I}^{t}}^{t}, \hat{\bm{O}}^{t}\bigr)}, \label{eq:object_loss_i} 
\end{align}
where $\hat{\bm{O}}^{t}$ denotes the ground-truth locations of group-relevant objects at $t$-th frame.
\section{Experiments}
\label{sec:exp}

%%%%%%%%%%%%%%%%%%%%%%%%%%%%%%%%%%%%%%%%%%%%%%%%%%

\subsection{Datasets}
\label{subsec:datasets}

%%%Two widely used public datasets are used.

\noindent\textbf{Volleyball dataset (VBD)~\cite{DBLP:conf/cvpr/VBD}:}
This dataset contains 4830 sequences extracted from 55 volleyball games. 4830 sequences are split into 3493 and 1337 sequences as the training and test sets, respectively. In each sequence, one of the predefined eight group activity labels (i.e., Left-spike, Right-spike, Left-set, Right-set, Left-pass, Right-pass, Left-winpoint, and Right-winpoint) is annotated manually.
%The action of each person is also annotated with a bounding box.

\noindent\textbf{NBA dataset (NBA)~\cite{DBLP:conf/eccv/NBA}:}
This dataset comprises 9172 sequences, which are split into 7624 training and 1548 test sequences, respectively.
%extracted from 181 basketball games.
%These 9172 sequences are split into 7624 training and 1548 test sequences, respectively.
One of the pre-defined nine group activity labels (i.e., 2p-succ, 2p-fail-off, 2p-fail-def, 2p-layup-succ, 2p-layup-fail-off, 2p-layup-fail-def, 3p-succ, 3p-fail-off, and 3p-fail-def) is annotated in each sequence.

%%%%%%%%%%%%%%%%%%%%%%%%%%%%%%%%%%%%%%%%%%%%%%%%%%

\subsection{Details}
\label{subsec:exp_det}

The image is resized to $288\times512$ for both VBD and NBA.
We train the entire network shown in Fig.~\ref{fig:overview_network} (except for the inpainting network) using Adam~\cite{DBLP:journals/corr/Adam} with cosine scheduling in two stages.
In the first stage, the flow loss, $\mathcal{L_{\mathrm{F}}}$, is used for 50 epochs with an initial learning rate of $5.0\times 10^{-5}$. In the second stage, the group-relevant object loss, $\mathcal{L_{\mathrm{O}}}$, is used for 30 epochs with an initial learning rate of $2.5\times 10^{-5}$.
%
%Following \cite{DBLP:conf/icml/Explora}, our method
%efficiently
%finetunes DINOv3 by training only the last two layers of its ViT.

While any object can be used as a group-relevant object, a ball is used in our experiments because it is always important for understanding team ball sports like volleyball and basketball. 
Such team sports are widely used as typical examples of group activities~\cite{DBLP:conf/cvpr/Dual-AI,DBLP:conf/cvpr/Detector_Free,DBLP:conf/eccv/flow_assist} in the literature.

For ball inpainting, LaMa~\cite{DBLP:conf/wacv/lama} is employed due to its high speed. While the center point of a ball in each frame (i.e., $\hat{\bm{O}}^{t}$) is annotated~\cite{DBLP:journals/pr/PerezLK22} in VBD and used for this ball inpainting, it is obtained by a ball tracker~\cite{DBLP:conf/bmvc/WASB} in NBA.

The flow values of each person are 
%provided from flow images
obtained by RAFT~\cite{DBLP:conf/eccv/raft}. The flow values of $p$-th person (i.e., $\hat{\bm{F}}^{p,t}$) are extracted from the center of this person's bounding box. 
%
%%% ..... (for space reduction) We compensate for camera panning by subtracting the median flow values in each frame.
% 
%In VBD, since the bounding box of each person is annotated~\cite{DBLP:conf/mva/SendoU19}, it is used to extract the flow values for that person. On the other hand, since there are no annotated bounding boxes in NBA, bounding boxes obtained by YOLOX~\cite{DBLP:journals/corr/YOLOX} are used.
While the bounding box of each person in each frame is annotated~\cite{DBLP:conf/cvpr/VBD} in VBD and used for this flow extraction, it is obtained by YOLOX~\cite{DBLP:journals/corr/YOLOX} in NBA.

%%% (described in Sec. 3.3.3) During training, to increase training capability using videos, the flows of all people and the ball locations are estimated in all frames with Eq.~\ref{eq:flow_vid}, \ref{eq:flow_img}, \ref{eq:obj_vid}, and \ref{eq:obj_img}.

\begin{table}[t]
    \centering
    \caption{Comparison with state-of-the-art self-supervised GAF learning methods~\cite{DBLP:conf/eccv/HRN,DBLP:conf/cvpr/GAFL} on VBD and NBA.
    While all scores of the other methods on VBD come from~\cite{DBLP:conf/cvpr/GAFL}, we obtained the scores on NBA with the codes and weights provided by~\cite{DBLP:conf/cvpr/GAFL} because NBA is not used in experiments shown in~\cite{DBLP:conf/cvpr/GAFL}.
    }
    \begin{tabular}{l|cc|cc}\hline
    & \multicolumn{2}{c|}{VBD}               & \multicolumn{2}{c}{NBA}                       \\ \hline
    Method        & Hit@1         & Hit@3         & Hit@1         & Hit@3         \\ \hline
    B1-Compact~\cite{DBLP:conf/eccv/HRN} & 30.3          & 59.9          & 14.9          & 39.5          \\
    B2-VGG19~\cite{DBLP:conf/eccv/HRN}      & 35.4          & 65.0          & 16.8          & 39.8          \\
    HRN~\cite{DBLP:conf/eccv/HRN}           & 31.2          & 57.6          & 15.5          & 37.1          \\
    %GAFL-ind~\cite{DBLP:conf/cvpr/GAFL}      & 55.0          & 79.2          & 23.9          & 49.2          \\
    GAFL~\cite{DBLP:conf/cvpr/GAFL}      & 61.1          & 82.4          & 24.7          & 50.4          \\ \hline\hline
    Ours          & \textbf{82.7} & \textbf{93.0} & \textbf{43.9} & \textbf{72.0} \\ \hline
    \end{tabular}
    \label{tab:comp_garet}
\end{table}

%%%%%%%%%%%%%%%%%%%%%%%%%%%%%%%%%%%%%%%%%%%%%%%%%%

\subsection{Evaluation in Group Activity Retrieval}
\label{subsec:garet}

Following~\cite{DBLP:conf/eccv/HRN,DBLP:conf/cvpr/GAFL}, our method is evaluated in the group activity retrieval task. Query videos are selected from the test set. For each query video, similar videos are retrieved from the training set. The similarity is defined as the Euclidean distance in the GAF space. The retrieval is regarded as a success if the group activity labels annotated in the query and the retrieved video are the same. In each retrieval, $K$ videos are retrieved, and Hit@K is evaluated, in accordance with the literature of information retrieval.
%%% ..... (for space reduction) Note that the annotated group activity labels are only used for the evaluation, while these labels are not used in training the network.

%%%%%%%%%%%%%%%%%%%%%%%%%%%%%%

\subsubsection{Comparative experiments}
\label{subsubsec:comp_exp}

Our method is compared with previous self-supervised GAF learning methods~\cite{DBLP:conf/eccv/HRN,DBLP:conf/cvpr/GAFL}.
Table~\ref{tab:comp_garet} shows the comparison on VBD and NBA. We can see that our method is the best in all metrics on VBD and NBA. Specifically, even compared with the second best, GAFL~\cite{DBLP:conf/cvpr/GAFL}, our method achieves a significant performance gain (i.e., $82.7-61.1=21.6$ and $43.9-24.7=19.2$ on VBD and NBA, respectively) in Hit@1. 
Since this gain is acquired by the main differences between GAFL and our method (i.e., the combination of DINOv3 and our proposed pretext tasks using $\mathcal{L}_{\mathrm{F}}$ and $\mathcal{L}_{\mathrm{O}}$), this gain demonstrates the effectiveness of the combination of DINOv3 and our pretext tasks.

\begin{figure}[t]
    \centering
    \includegraphics[width=\columnwidth]{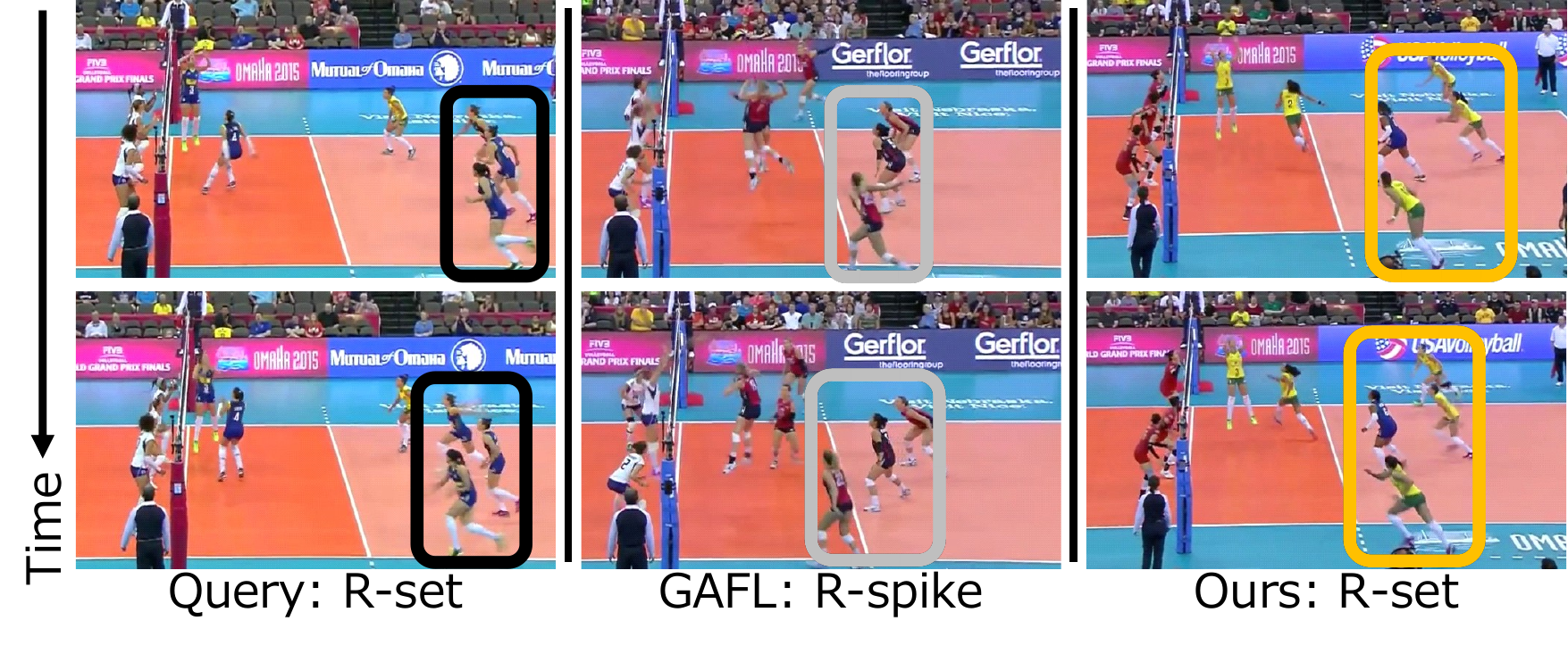}\\
    \vspace*{-2mm}
    (a) In the case of an R-set query.\\
    \vspace*{1mm}
    \includegraphics[width=\columnwidth]{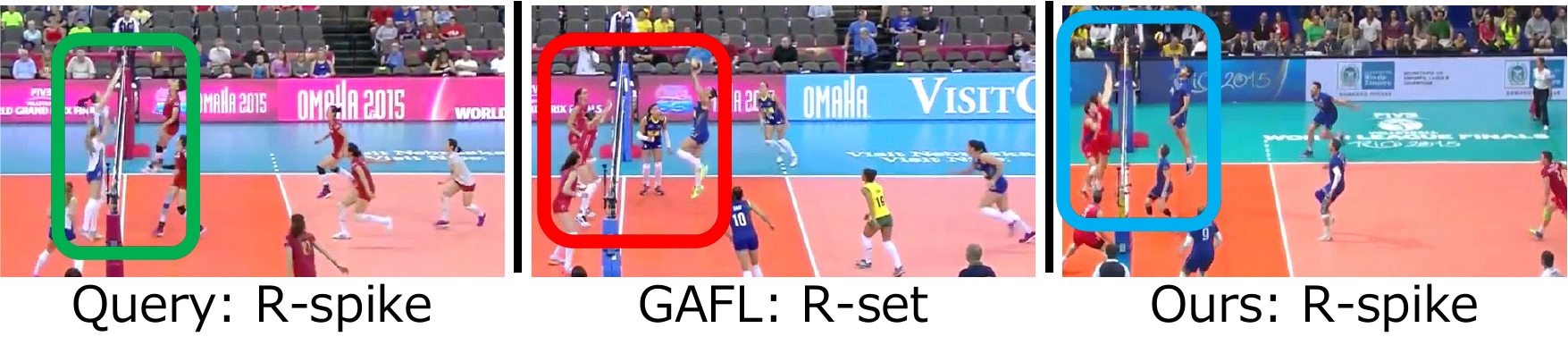}\\
    \vspace*{-2mm}
    (b) In the case of an R-spike query.\\
    \vspace*{-2mm}
    \caption{Visual comparison of group activity retrieval on VBD. (a) R-set query. (b) R-spike query.
    }
    \label{fig:ret_comp_vbd}
\end{figure}

The results of group activity retrieval on VBD are visualized in Fig.~\ref{fig:ret_comp_vbd}.
%%% .... can be omitted
In (a), for the R-set query, our method correctly retrieves an R-set video, whereas GAFL retrieves an R-spike video. 
In the query video of (a), the players in the black-outlined regions move left for R-set. Our method successfully retrieves the video of R-set in which the players in the yellow-outlined regions also move left. In contrast, GAFL retrieves the video of R-spike in which the players in the grey-outlined regions remain still. This suggests that our method captures the dynamics of the players, such as for a set activity.

%In (b), for an R-spike query, our method correctly retrieves R-spike, whereas GAFL retrieves R-set. 
In the query video of (b), the jumping spiker and the jumping blockers are observed in the green-outlined region for R-spike. Our method successfully retrieves an R-spike video that contains an interaction between the jumping spiker and the jumping blockers in the blue-outlined region. In contrast, GAFL retrieves the video of R-set, in which some players gather near the net, while no one is jumping in the red-outlined region.
This is because GAFL retrieves this R-set video based solely on the local appearance similarity between the players in the green- and red-outlined regions.
%because GAFL is trained only with a pretext task with appearance reconstruction.
On the other hand, our method finds the difference between the local dynamics features of jumping people in R-spike and standing people in R-set.

\begin{figure}[t]
    \centering
    \includegraphics[width=\columnwidth]{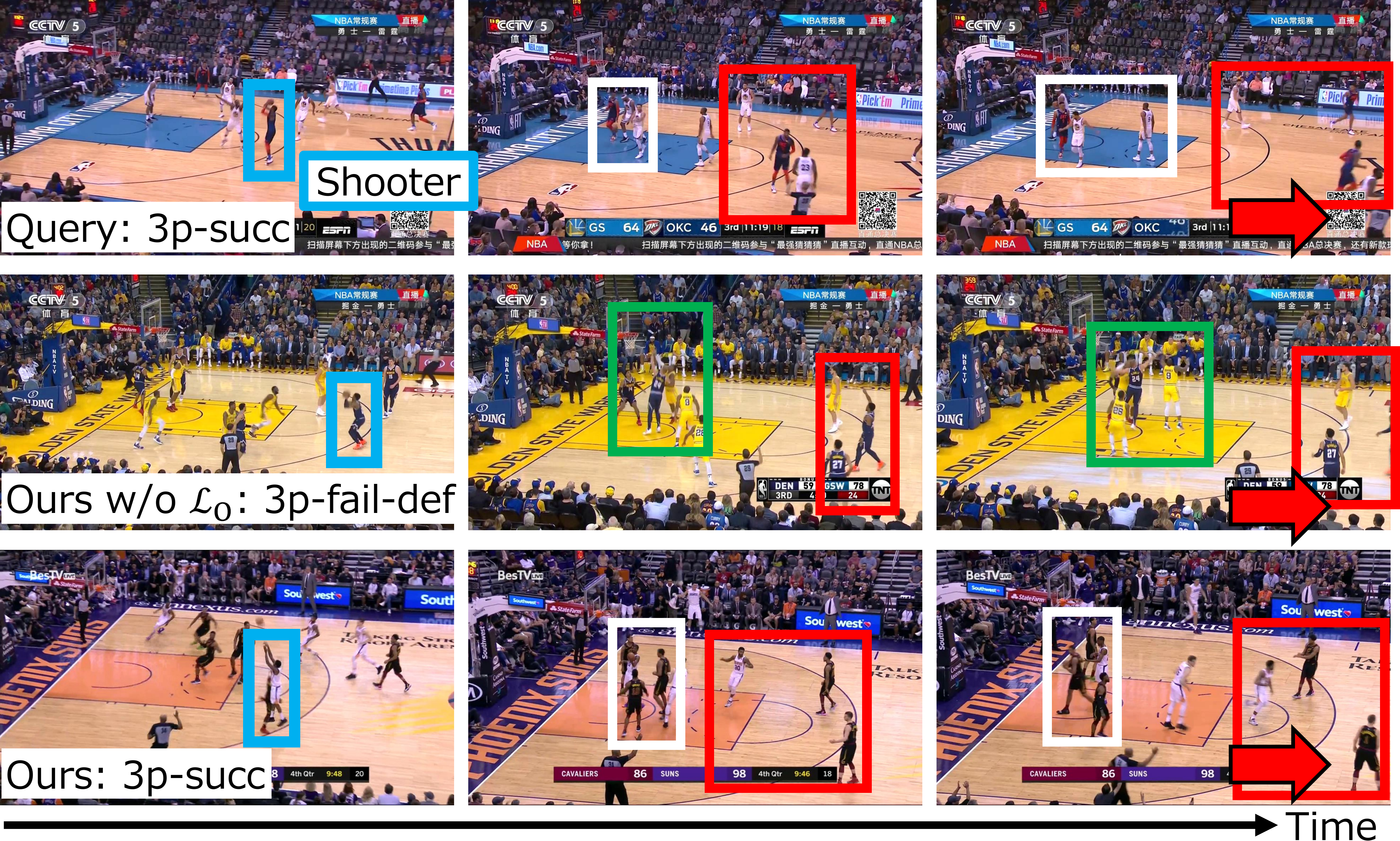}\\
    \vspace*{-2mm}
     \caption{
     Visual comparison of group activity retrieval on NBA when the query video is a 3p-succ video.
     }
     \label{fig:ret_comp_nba_ours}
\end{figure}

\begin{table}[t]
    \centering
    \caption{Ablation studies on VBD and NBA.
    $\color{mygreen}\mathrm{\checkmark}$ in (d) and (i) indicates that $\mathcal{L}_{\mathrm{F}}$ and $\mathcal{L}_{\mathrm{O}}$ are used simultaneously. \textcolor{red}{$\mathrm{\checkmark}$} and \textcolor{cyan}{$\mathrm{\checkmark}$} indicate a loss used in the first and second stages, respectively.
    }
    \vspace*{-3mm}
    \begin{tabular}{c|cc|cc|c|c}
    \hline
        & \multicolumn{2}{c|}{$\mathcal{L}_{\mathrm{F}}$}                                              & \multicolumn{2}{c|}{$\mathcal{L}_{\mathrm{O}}$}                                              & VBD           & NBA           \\ \cline{2-7} 
        & \multicolumn{1}{c|}{$\mathcal{L}_{\mathrm{F}, \bm{G}}$} & $\mathcal{L}_{\mathrm{F}, \bm{I}}$ & \multicolumn{1}{c|}{$\mathcal{L}_{\mathrm{O}, \bm{G}}$} & $\mathcal{L}_{\mathrm{O}, \bm{I}}$ & Hit@1         & Hit@1         \\ \hline
    (a) & \multicolumn{2}{c|}{-}                                                                       & \multicolumn{2}{c|}{-}                                                                       & 43.0          & 22.6          \\ \hline
    (b) & \multicolumn{2}{c|}{$\mathrm{\checkmark}$}                                                   & \multicolumn{2}{c|}{-}                                                                       & 75.4          & 34.1          \\ \hline
    (c) & \multicolumn{2}{c|}{-}                                                                       & \multicolumn{2}{c|}{$\mathrm{\checkmark}$}                                                   & 74.0          & 40.3          \\ \hline
    (d) & \multicolumn{2}{c|}{$\color{mygreen}\mathrm{\checkmark}$($w_{\mathrm{F}}$=1)}                               & \multicolumn{2}{c|}{$\color{mygreen}\mathrm{\checkmark}$($w_{\mathrm{O}}$=1)}                               & 77.9          & 42.9          \\ \hline
    (e) & \multicolumn{2}{c|}{$\color{red}\mathrm{\checkmark}$}                                        & \multicolumn{2}{c|}{$\color{cyan}\mathrm{\checkmark}$}                                       & \textbf{82.7} & 43.9          \\ \hline
    (f) & \multicolumn{2}{c|}{$\color{cyan}\mathrm{\checkmark}$}                                       & \multicolumn{2}{c|}{$\color{red}\mathrm{\checkmark}$}                                        & 76.7          & 36.3          \\ \hline
    (g) & \multicolumn{1}{c|}{$\color{red}\mathrm{\checkmark}$}   & -                                  & \multicolumn{2}{c|}{$\color{cyan}\mathrm{\checkmark}$}                                       & 80.5          & 43.2          \\ \hline
    (h) & \multicolumn{2}{c|}{$\color{red}\mathrm{\checkmark}$}                                        & \multicolumn{1}{c|}{$\color{cyan}\mathrm{\checkmark}$}  & -                                  & 80.6          & 43.7          \\ \hline
    (i) & \multicolumn{2}{c|}{$\color{mygreen}\mathrm{\checkmark}$($w_{\mathrm{F}}$=1)}                               & \multicolumn{2}{c|}{$\color{mygreen}\mathrm{\checkmark}$($w_{\mathrm{O}}$=3)}                               & 81.2          & \textbf{46.3}          \\ \hline
    %(j) & \multicolumn{2}{c|}{$\mathrm{\checkmark}$($w_{\mathrm{F}}$=1)}                               & \multicolumn{2}{c|}{$\mathrm{\checkmark}$($w_{\mathrm{O}}$=5)}                               & 81.0          & \textbf{47.0} \\ \hline
    %(k) & \multicolumn{2}{c|}{$\mathrm{\checkmark}$($w_{\mathrm{F}}$=1)}                               & \multicolumn{2}{c|}{$\mathrm{\checkmark}$($w_{\mathrm{O}}$=10)}                              & 81.7          & 46.7          \\ \hline
    \end{tabular}
    \label{tab:abl_vbd}
    \vspace*{-2mm}
\end{table}

%%%%%%%%%%%%%%%%%%%%%%%%%%%%%%

\subsubsection{Ablation studies on Loss Functions}
\label{subsubsec:abl_garet}

The effects of our proposed loss functions (i.e., $\mathcal{L}_{\mathrm{F}}$ and $\mathcal{L}_{\mathrm{O}}$) are validated by ablation studies shown in Table~\ref{tab:abl_vbd}. 
In (a), neither $\mathcal{L}_{\mathrm{F}}$ nor $\mathcal{L}_{\mathrm{O}}$ is used.
Since no training is possible with no loss functions, the pretrained DINOv3 as-is is used to extract CLS tokens, and they are aggregated via temporal max pooling in (a). With the exception of (a), all results are obtained with our proposed network shown in Fig.~\ref{fig:overview_network}.

Only $\mathcal{L}_{\mathrm{F}}$ and $\mathcal{L}_{\mathrm{O}}$ are used in (b) and (c), respectively.
The performance is improved by $\mathcal{L}_{\mathrm{F}}$; Hit@1 is improved to 75.4 from 43.0 on VBD, and to 34.1 from 22.6 on NBA.
With $\mathcal{L}_{\mathrm{O}}$, Hit@1 is improved to 74.0 from 43.0 on VBD, and to 40.3 from 22.6 on NBA.
With these results, we conclude that both $\mathcal{L}_{\mathrm{F}}$ and $\mathcal{L}_{\mathrm{O}}$ contribute to GAF learning.

% ..... Old scenario: Learning with Lo is more difficult than learning with Lf.
%In (d), (e), and (f), the impact of the joint learning strategy is verified. The best scores are acquired in (e). We can interpret this result as stemming from the difficulty of learning our proposed network using $\mathcal{L}_{\mathrm{O}}$, while $\mathcal{L}_{\mathrm{O}}$ can contribute to the generalization ability of the network. One difficulty is that ....
%
The impact of jointly learning $\mathcal{L}_{\mathrm{F}}$ and $\mathcal{L}_{\mathrm{O}}$ is demonstrated in (d), (e), and (f).
In (d), $\mathcal{L}_{\mathrm{F}}$ and $\mathcal{L}_{\mathrm{O}}$ are used simultaneously.
%
%In (e) and (f), on the other hand, $\mathcal{L}_{\mathrm{F}}$ and $\mathcal{L}_{\mathrm{O}}$ are used in the first training stage, and then $\mathcal{L}_{\mathrm{O}}$ and $\mathcal{L}_{\mathrm{F}}$ are used in the second stage, respectively.
In (e) and (f), on the other hand, either $\mathcal{L}_{\mathrm{F}}$ or $\mathcal{L}_{\mathrm{O}}$ is used in each of the first and second stages. In (e), $\mathcal{L}_{\mathrm{F}}$ and $\mathcal{L}_{\mathrm{O}}$ are used in the first and second stages, respectively. In (f), the order is reversed.
Of (d), (e), and (f), (e) achieves the best score.
We can interpret the superiority of (e) as stemming from the higher impact of global features on group activity representation. That is, while $\mathcal{L}_{\mathrm{F}}$ also contributes to GAF learning so that (d), (e), and (f) are slightly better than (c) in most cases, $\mathcal{L}_{\mathrm{F}}$ interferes with the global feature learning using $\mathcal{L}_{\mathrm{O}}$ in (d) and (f).
In (e), on the other hand, $\mathcal{L}_{\mathrm{O}}$ has a greater impact on the final GAF space than $\mathcal{L}_{\mathrm{F}}$ because $\mathcal{L}_{\mathrm{F}}$ is used only in the first stage, and then $\mathcal{L}_{\mathrm{O}}$ updates the GAF space in the second stage.
To support this interpretation, we increase the weight of  $\mathcal{L}_{\mathrm{O}}$ when $\mathcal{L}_{\mathrm{F}}$ and $\mathcal{L}_{\mathrm{O}}$ are used simultaneously.
$\mathcal{L}_{\mathrm{O}}$ is multiplied by a weight denoted by $w_{\mathrm{O}}$.
As shown in (i), where $w_{\mathrm{O}} = 3$, a larger weight for $\mathcal{L}_{\mathrm{O}}$ improves the scores compared with (d) and (f).
This result supports our interpretation.
%%% ..... future work: curriculum learning is useful?

In addition, our proposed auxiliary loss functions, $\mathcal{L}_{\mathrm{F}, \bm{I}}$ and $\mathcal{L}_{\mathrm{O}, \bm{I}}$, are ablated in (g) and (h), respectively. 
In the comparison of (g) and (h) with (e), (e) is better than (g) and (h), demonstrating the effectiveness of $\mathcal{L}_{\mathrm{F}, \bm{I}}$ and $\mathcal{L}_{\mathrm{O}, \bm{I}}$. This result suggests that $\mathcal{L}_{\mathrm{F}, \bm{I}}$ and $\mathcal{L}_{\mathrm{O}, \bm{I}}$ encourage DINOv3 to learn local dynamics features and global group activity representation, respectively, because $\mathcal{L}_{\mathrm{F}, \bm{I}}$ and $\mathcal{L}_{\mathrm{O}, \bm{I}}$ train DINOv3 only, leaving the GAF learning network.

To further demonstrate the significance of $\mathcal{L}_{\mathrm{O}}$, Fig.~\ref{fig:ret_comp_nba_ours} shows visual examples.
For a 3p-succ query, our method correctly retrieves a 3p-succ video, whereas our method without $\mathcal{L}_{\mathrm{O}}$ retrieves a 3p-fail-def video. These two classes are similar because both involve a 3-point shot followed by a defensive possession (e.g., several players move to the right after the shot, as shown by the players enclosed in the red boxes in Fig.~\ref{fig:ret_comp_nba_ours}), while these two classes can be classified based on how other players behave depending on the shot's success or failure. This classification requires the global context among multiple players. Specifically, this 3p-succ query includes a frame where the other players do not move 
%aggressively
(as enclosed by the white boxes in Fig.~\ref{fig:ret_comp_nba_ours}) because the play ends once. On the other hand, several players, enclosed within the green boxes, keep making strategic moves globally in 3p-fail-def videos.
%
%However, our method without $\mathcal{L}_{\mathrm{O}}$ focuses only on local cues such as the players moving left to right and thus fails to distinguish between 3p-succ and 3p-fail-def videos, in both of which the players locally move left to right.
However, such global spatial relations are not learned in our method without $\mathcal{L}_{\mathrm{O}}$.
%, which fails to distinguish between 3p-succ and 3p-fail-def videos.
%
On the other hand, by using $\mathcal{L}_{\mathrm{O}}$ with $\mathcal{L}_{\mathrm{F}}$ in our method, $\mathcal{L}_{\mathrm{F}}$ encourages the GAF space to increase the similarity between 3p-succ and 3p-fail-def videos, and $\mathcal{L}_{\mathrm{O}}$ allows for discriminating between the global contexts between 3p-succ and 3p-fail-def videos in the second training stage.

%%%%%%%%%%%%%%%%%%%%%%%%%%%%%%

\subsubsection{Ablation studies on Inpainting}
\label{subsubsec:abl_inpaint_garet}

\if 0
\begin{table}[t]
    \centering
    \caption{Detailed analysis of inpainting on VBD and NBA.}
    \begin{tabular}{l|c|c}
    \hline
    \multirow{2}{*}{Input image} & VBD           & NBA           \\ \cline{2-3}
    & Hit@1         & Hit@1         \\ \hline
    Original          & 81.6          & 35.9          \\
    Masking           & 69.3          & 26.7          \\ \hline\hline
    Inpainting (Ours) & \textbf{82.7} & \textbf{43.9} \\ \hline
    \end{tabular}
    \label{tab:mask_comp_vbd}
\end{table}
\fi

\begin{table}[t]
    \centering
    \caption{Detailed analysis of inpainting on VBD and NBA.}
    \vspace*{-1mm}
    \begin{tabular}{l|cc|cc}\hline
    \multirow{2}{*}{Input image} & \multicolumn{2}{c|}{VBD}               & \multicolumn{2}{c}{NBA}                       \\ \cline{2-5}
    & Hit@1         & Hit@3         & Hit@1         & Hit@3         \\ \hline
    Original          & 81.6 & 92.0          & 35.9 & 65.3          \\
    Masking           & 69.3 & 87.1          & 26.7 & 54.1        \\ \hline\hline
    Inpainting (Ours) & \textbf{82.7} & \textbf{93.0} & \textbf{43.9} & \textbf{72.0} \\ \hline
    \end{tabular}
    \label{tab:mask_comp_vbd}
\end{table}

Table~\ref{tab:mask_comp_vbd} shows the effectiveness of object inpainting. In ``Original,'' the original images of an input video are directly fed into DINOv3 without ball inpainting. In ``Masking,'' the region of a ball is filled with black pixels to remove the appearance of the ball. In ``Inpainting (Ours),'' which is our proposed method, the region of a ball is inpainted.

In comparison with ``Original,'' our method is 1.1 ($82.7-81.6$) better on VBD and 8.0 ($43.9-35.9$) better on NBA.
%These results validate that inpainting encourages learning of scene context in GAFs.
%
On the other hand, we observe a significant performance drop in ``Masking.'' This is because the unique appearance cues of the masked region (i.e., the region filled with black pixels) are easy to localize, thereby interfering with the learning of global context in GAFs.

%%% (Due to the space limitation, this paragraph is deleted.)
%We next explain why there are differences in the results between VBD and NBA. Unlike NBA, the group activity classes on VBD are separated depending on the side of the court where the play occurs (e.g., Left-spike and Right-spike). Learning the ball location itself leads to a moderate improvement in performance. As a result, the performance gap compared with using the original images as input is small. In contrast, the NBA labels specify shot type (i.e., 2p, 2p-layup, 3p) and outcome (i.e., success, fail-off, fail-def), not the side of the court. This requires understanding scene context, including how multiple players interact, beyond just the ball position. Consequently, compared with training that uses original images as input, our method achieves a substantial performance improvement on NBA.

\begin{table}[t]
    \centering
    \caption{
    Different feature extractors are used in our proposed network. Due to image feature extractor constraints, we use $256\times256$ inputs for ``SigLIP 2~\cite{SigLIP2}'' and $224\times224$ for the others.
    }
    \vspace*{-1mm}
    \begin{tabular}{lc|c|c}
    \hline
    \multicolumn{2}{l|}{Dataset}                          & VBD           & NBA           \\ \hline
    \multicolumn{1}{l|}{Feature extractor} & Architecture & Hit@1         & Hit@1         \\ \hline
    \multicolumn{1}{l|}{Supervised ViT~\cite{DBLP:conf/iclr/supervised_ViT}}    & ViT-L/16     & 41.9          & 29.8          \\
    \multicolumn{1}{l|}{MAE~\cite{DBLP:conf/cvpr/MAE}}               & ViT-L/16     & 55.4          & 40.1          \\
    \multicolumn{1}{l|}{CLIP~\cite{DBLP:conf/icml/CLIP}}              & ViT-L/14     & 59.8          & 37.9          \\
    \multicolumn{1}{l|}{SigLIP 2~\cite{SigLIP2}}          & ViT-L/16     & 62.2          & 38.1          \\ \hline\hline
    \multicolumn{1}{l|}{DINOv3~\cite{DINOv3} (Ours)}     & ViT-L/16     & \textbf{73.1} & \textbf{40.3} \\ \hline
    \end{tabular}
    \label{tab:backbone_comp}
\end{table}

%%%%%%%%%%%%%%%%%%%%%%%%%%%%%%

\subsubsection{Effects of Image Feature Extractors}

Table~\ref{tab:backbone_comp} shows the results with different pre-trained image feature extractors in our method. 
DINOv3~\cite{DINOv3} achieves the best score, Hit@1 = 73.1 and 40.3 on VBD and NBA, respectively.
The superiority of DINOv3 over Supervised ViT~\cite{DBLP:conf/iclr/supervised_ViT}, MAE~\cite{DBLP:conf/cvpr/MAE}, and CLIP~\cite{DBLP:conf/icml/CLIP} can be interpreted as alignment of local and global features.
%, both of which are important for group activity representation.
%
SigLIP 2~\cite{SigLIP2}, which is an image-text encoder, is trained in a DINO-like manner to align local and global views.
However, SigLIP 2 may lack the ability to learn the detailed local features due to the brief nature of image-text alignment, as verified in the image segmentation task~\cite{DINOv3}.

%%%%%%%%%%%%%%%%%%%%%%%%%%%%%%%%%%%%%%%%%%%%%%%%%%

\subsection{Evaluation in Group Activity Recognition}
\label{subsec:gar}

\begin{figure}[t]
    \centering
    \includegraphics[width=\columnwidth]{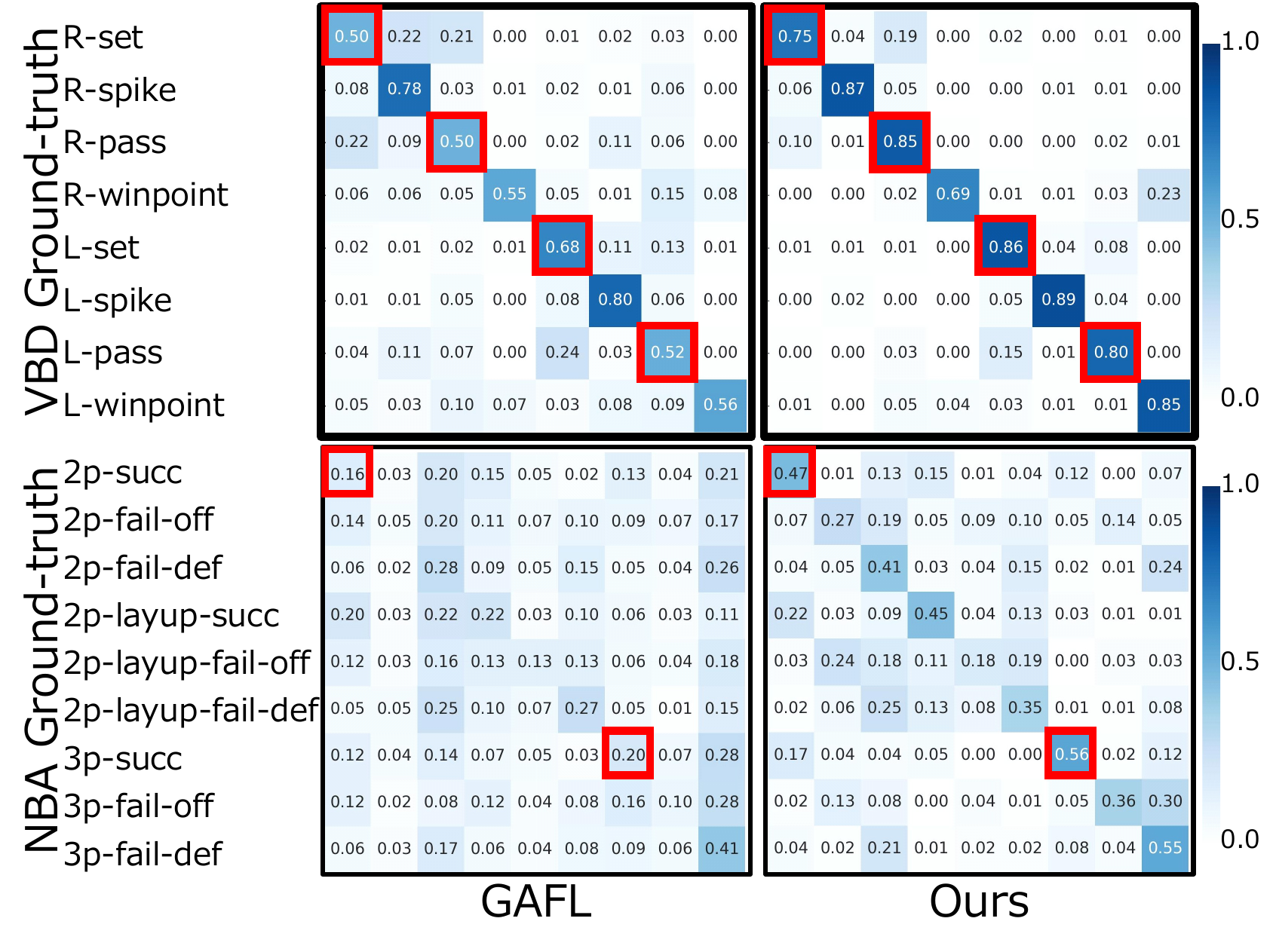}
    \vspace*{-7mm}
    \caption{Confusion matrices of GAR by nearest neighbor retrieval on VBD and NBA. Each row and column shows the ground-truth and recognized group activity, respectively.}
    \label{fig:cm_nn_ret_vbd}
    \vspace*{-2mm}
\end{figure}

%%%%%%%%%%%%%%%%%%%%%%%%%%%%%%

\subsubsection{Nearest Neighbor Retrieval}
\label{subsubsec:nn_ret}

Our network is trained to extract GAFs, but is not trained to recognize group activity classes. However, we can use the GAF space to classify each query video into one of the 
%group activity
classes via nearest neighbor retrieval if a group activity label is provided for each video in a retrieval database, which is replaced with the training set in our experiments.

\begin{figure}[t]
    \centering
    \includegraphics[width=\columnwidth]{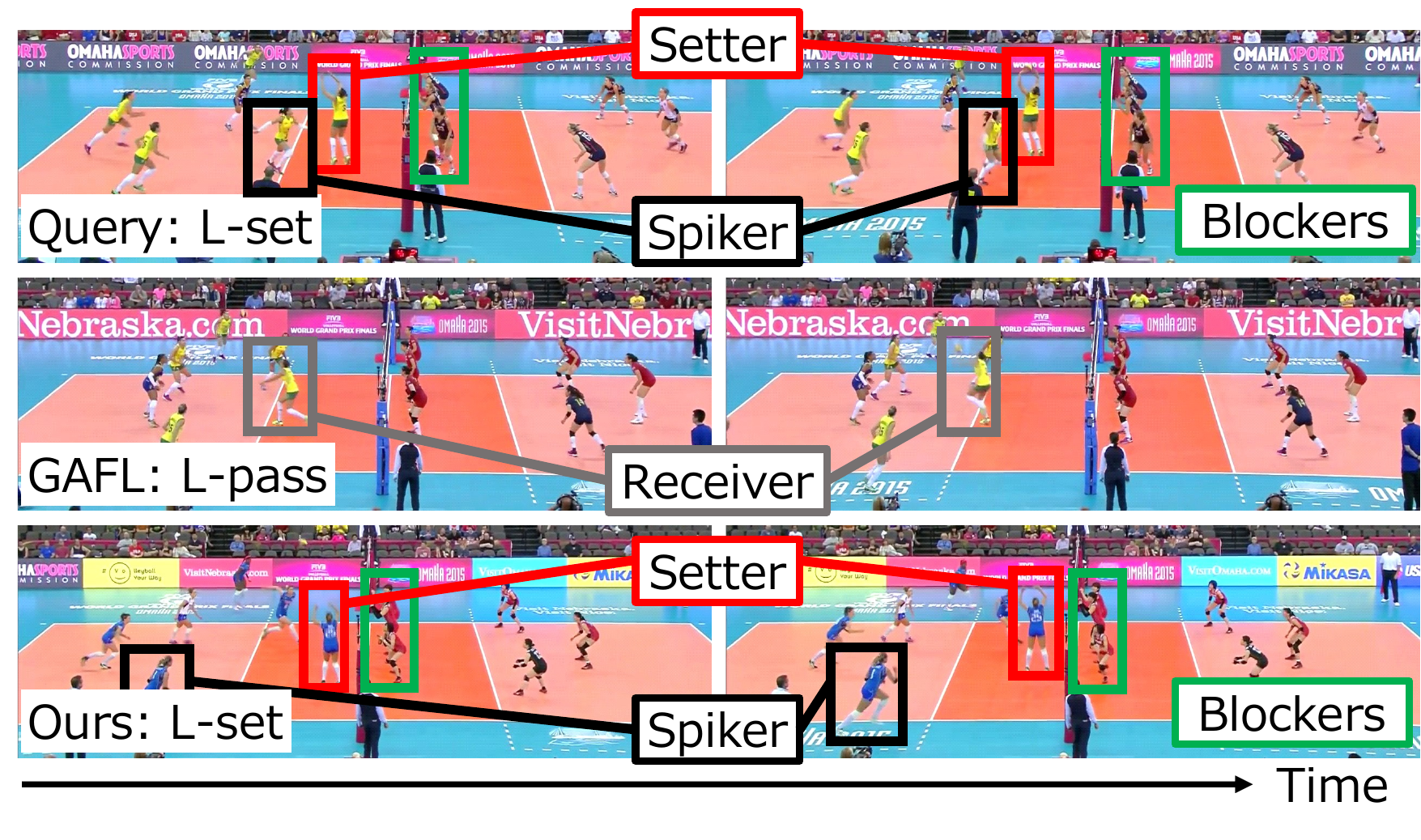}
    \vspace{-8mm}
    \caption{Visual comparison of group activity retrieval on VBD.
     }
    \label{fig:vis_VBD_gafl_ours}
\end{figure}

\noindent{\bf VBD.}
As shown in the top of Fig.~\ref{fig:cm_nn_ret_vbd}, on VBD, our method surpasses GAFL across all group activity classes, with especially clear gains for R-set, R-pass, and L-set, L-pass, as highlighted by the red box cells.
GAFL often retrieves set videos even when the query is a pass video (and vice versa). This issue can be explained as follows.
Set and pass videos are similar because, in both classes, a player receives a ball, as indicated by ``Setter'' in the L-set videos and ``Receiver'' in the L-pass video of Fig.~\ref{fig:vis_VBD_gafl_ours}. However, they are contextually different. A setter plays with teammates nearby who are ready to spike (``Spiker'' in Fig.~\ref{fig:vis_VBD_gafl_ours}) and players on the opposing side preparing to defend (``Blockers'' in Fig.~\ref{fig:vis_VBD_gafl_ours}). In contrast, a pass video should include a receiver but no spikers or setters.
While the local dynamics features of each player allow us to identify the role of the player (e.g., a setter or a receiver), the global spatial configuration of the players is not represented by these local features.
In contrast, $\mathcal{L}_{\mathrm{O}}$ enables GAFs to discriminate among these global contexts.

\begin{figure}[t]
    \centering
    \includegraphics[width=\columnwidth]{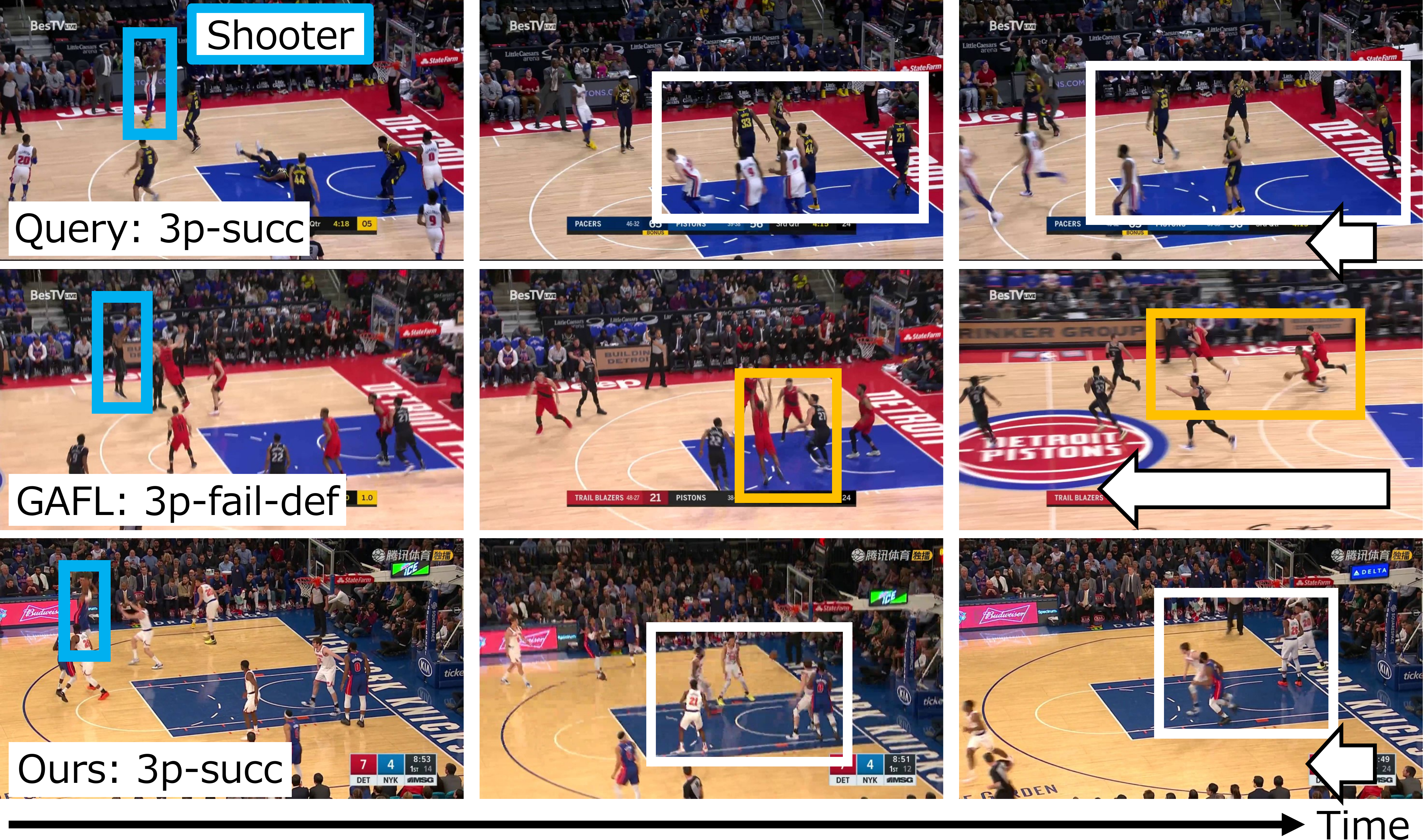}
    \vspace{-6mm}
    \caption{Visual comparison of group activity retrieval on NBA.
     }
    \label{fig:vis_nba_gafl_ours}
    \vspace*{-4mm}
\end{figure}

\noindent{\bf NBA.}
As shown in the bottom of Fig.~\ref{fig:cm_nn_ret_vbd}, our method also surpasses GAFL on NBA across all group activity classes, with especially clear gains for 2p-succ and 3p-succ, as highlighted by the red box cells.
This is because GAFL often misclassifies 2p/3p-succ and 2p/3p-fail-def.
%%%We also observe that GAFL often retrieves 3p-succ even when the query is 3p-fail-def (and vice versa).
These results are visualized in Fig.~\ref{fig:vis_nba_gafl_ours}.
Both succ and fail-def are similar in that each player transitions to a defensive phase after a shot, as shown by the players enclosed in the white and yellow rectangles in Fig.~\ref{fig:vis_nba_gafl_ours}.
However, from the viewpoint of a global context, the play briefly stops after the shot in succ; players enclosed by the white rectangles do not move much. In fail-def, by contrast, some players contend for the ball; the players enclosed by the yellow rectangles move from right to left. Thus, the two classes differ in how multiple players interact with the ball after the shot.
Such people-object interactions are better represented in our method.

\begin{table}[t]
    \centering
    \caption{
    Comparison with supervised group activity recognition methods on VBD and NBA. The results of all methods are copied from their papers. The four methods above the middle line take the whole image as input, while the five methods between the middle line and ours use person bounding boxes together with the whole image in both training and inference. For a fair comparison, in all methods, the only manually annotated labels are group activity classes, and only images are used in inference.
    }
    \vspace*{-2mm}
    \begin{tabular}{lc|c|c}
    \hline
    Method        & Extractor & VBD           & NBA \\ \hline

    DFWSGAR~\cite{DBLP:conf/cvpr/Detector_Free}       & ResNet-18                                                   & 90.5          & 75.8                                                 \\
    SOGAR~\cite{DBLP:journals/access/SOGAR}         & ViT-Base                                                    & 93.1          & \textbf{83.3}                                                 \\
    Flaming-Net~\cite{DBLP:conf/eccv/flow_assist}   & Inception-v3                                                & 93.3          & 79.1                                               \\
    LiGAR~\cite{DBLP:conf/wacv/LiGAR}   & ResNet-18                                                   & 74.8          & 62.7                                                 \\ \hline
    SAM~\cite{DBLP:conf/eccv/NBA}           & ResNet-18                                                   & 86.3          & 54.3                                                 \\
    Dual-AI~\cite{DBLP:conf/cvpr/Dual-AI} & Inception-v3                                                & -             & 58.1                                                 \\
    KRGFormer~\cite{DBLP:journals/tcsv/KRGFormer}     & Inception-v3                                                & 92.4          & 72.4                                                 \\
    MP-GCN~\cite{DBLP:conf/eccv/skeleton-based}        & YOLOV8x                                                     & 92.8          & 78.7                                        \\
    FAGAR~\cite{DBLP:journals/pr/WangLLZGFC25}        & YOLOV8x                                                     & 85.2          & -                                        \\ \hline\hline
    Ours          & DINOv3                                                      & \textbf{93.9} & 73.0                                        \\ \hline
    \end{tabular}
    \label{tab:comp_wsgar}
\end{table}

%%%%%%%%%%%%%%%%%%%%%%%%%%%%%%

\subsubsection{Fine-tuning for Supervised Recognition}

% Our proposed network, shown in Fig.~\ref{fig:overview_network}, 
Our GAF learning can be used not only for recognition by nearest neighbor retrieval (Sec.~\ref{subsubsec:nn_ret}), but also as a pretraining for supervised recognition.
% This supervised recognition is implemented so that a GAF is fed into a one-layer linear classifier, and the whole network, including this classifier and our proposed network, is trained with group activity class supervision.
% 
In this supervised recognition, our network, followed by a one-layer linear classifier, is fine-tuned with group activity class supervision.
% This supervised recognition is implemented using our GAF network, followed by a one-layer linear classifier. The whole network is trained with group activity supervision.

\if 0
% move to supp
The recognition results are shown in Table~\ref{tab:comp_wsgar}. 
We use Multi-class Classification Accuracy (MCA) and Merged MCA~\cite{DBLP:conf/eccv/NBA}, in which the pass and set classes are merged into one class, as evaluation metrics. 
% 
Our method is the best in both MCA and Merged MCA. These results demonstrate that our proposed GAF learning is also effective as pre-training for supervised group activity recognition.
\fi
%
% The recognition results are shown in Table~\ref{tab:comp_wsgar}.
% 
% We use Multi-class Classification Accuracy (MCA).
Table~\ref{tab:comp_wsgar} shows the recognition results evaluated with Multi-class Classification Accuracy (MCA). 
Our method performs best on VBD.
This result validates that our GAF learning is effective as pre-training for supervised recognition.
%
% Although it does not achieve the best performance on NBA, it still yields competitive performance, supporting the effectiveness of our pretraining. 
% 
% A detailed discussion of the possible reasons for the relatively lower performance on NBA is provided in Sec.~\textcolor{cvprblue}{7.2.2} of the Supp.
% 
% On NBA, while our method achieves comparable performance compared with other methods, SOGAR (83.3\%) is better than our method (73.0\%) to a certain degree. 
On NBA, our method (73.0\%) is competitive, yet SOGAR (83.3\%) shows superior performance.
%
% The performance gap is likely due to our temporal max-pooling, which over-compresses long sequences and hinders the discrimination of similar group activities. Unlike NBA-specialized designs~[\textcolor{cvprblue}{6, 26, 29}], our generic architecture lacks explicit long-term temporal modeling. Enhancing GAFs with more flexible aggregation modules remains a promising direction for future work while preserving our model's simplicity.
% 
The performance gap likely stems from temporal max-pooling over-compressing long sequences, whereas NBA-specialized methods~[\textcolor{cvprblue}{6, 26, 29}] utilize richer temporal modeling.
See the details in the Supp. Sec.~\textcolor{cvprblue}{7.2.2}.
We leave flexible aggregation for future work to preserve simplicity. 
%
% This performance gap may be addressed by long-term modeling specialized for long NBA sequences. See the details in the Supp. Sec.~\textcolor{cvprblue}{7.2.2}.
% 
% The performance gap can be closed by incorporating temporal modeling specialized for the temporal sequence of the NBA as used in SOGAR. A detailed discussion about the richer temporal modeling is provided in Sec.~\textcolor{cvprblue}{7.2.2} of the Supp.

%%%%%%%%%%%%%%%%%%%%%%%%%%%%%%%%%%%%%%%%%%%%%%%%%%

\subsection{Visualization of GAFs}
\label{subsec:vis}

\begin{figure}[t]
    \centering
    \includegraphics[width=\columnwidth]{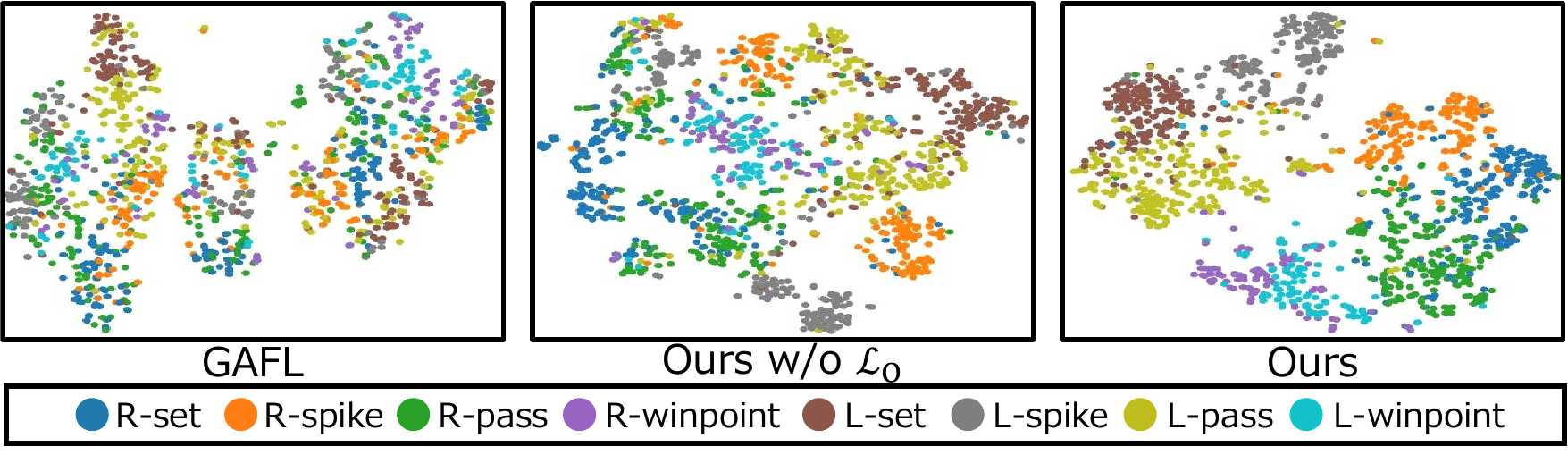}
    \vspace*{-6mm}
    \caption{Visualization of GAF spaces on VBD. GAFs extracted from the test set are projected to 2D by t-SNE~\cite{JMLR:v9:vandermaaten08a}. The color of each sample indicates its ground-truth class. %Ours w/o $\mathcal{L}_{\mathrm{O}}$ is trained only with the person flow estimation task ($\mathcal{L}_{\mathrm{F}}$). Ours (i.e., complete proposed method) is in addition to $\mathcal{L}_{\mathrm{F}}$, trained with $\mathcal{L}_{\mathrm{O}}$.
     }
     \label{fig:vis_gaf_space}
     \vspace*{-2mm}
\end{figure}

GAF spaces are visualized in Fig.~\ref{fig:vis_gaf_space}. 
This visualization reveals that, even when trained with $\mathcal{L}_{\mathrm{F}}$ alone, samples within the same class form compact clusters in the GAF space, while different classes are more mixed
in GAFL~\cite{DBLP:conf/cvpr/GAFL}. In our method, adding $\mathcal{L}_{\mathrm{O}}$ enables GAFs to learn global scene context, which further reduces the inner-class variance.
\section{Concluding Remarks}
\label{sec:conclusion}

% %
% Future work Future work Future work Future work. Future work Future work Future work Future work. Future work Future work Future work Future work. Future work Future work Future work Future work. Future work Future work. 

%\noindent{\bf Conclusion.}
This paper proposed group activity feature learning without group activity annotations by adapting DINOv3 to both local dynamics and global context features, which are essential for representing group activities. 
%Using person flow estimation and group-relevant object location estimation as complementary pretext tasks, our method learns local-dynamics- and global-group-aware features, which are essential for representing group activities.
Our method learns these features through two pretext tasks: person flow estimation and group-relevant object location estimation.
%Experiments on public datasets demonstrated state-of-the-art results in group activity retrieval and recognition, and ablations validated the role of each component. 
%
% We hope this work encourages future foundation-model-based approaches to understanding group activity, which reduce annotation costs and alleviate the need to predefine complex or even unknown group activity classes, while remaining easy to integrate into broader multi-person perception systems.

\noindent{\bf Limitations and Future Work.}
%\noindent{\bf Future Work.}
% Our proposed method still relies on pseudo-labels from external modules (e.g., flow estimators and object detectors), and it has been validated primarily in ball-centric sports scenarios. Future work includes reducing this reliance and extending the object-related prediction to multiple activity-relevant elements (e.g., a ball, goal, or court key areas) as well as more abstract elements (e.g., interaction hotspots such as entrances, ticket gates, or reception counters in social scenes), so that the framework can cover broader multi-person domains.
%
As with many group activity recognition methods~\cite{DBLP:conf/eccv/NBA,DBLP:conf/cvpr/Detector_Free,DBLP:conf/cvpr/ChappaNNSLDL23,DBLP:conf/eccv/flow_assist}, our method is validated only in ball-centric sports scenarios. Future work includes more experiments and extensions to other sport-related objects (e.g., the ball, the goal, and key court areas) as well as more generic objects
%(e.g., interaction hotspots such as entrances, ticket gates, or reception desks in social scenes), so that our method can cover broader scenarios.
(e.g., interaction hotspots such as entrances and reception desks), so that our method can cover broader scenarios.
{
    \small
    \bibliographystyle{ieeenat_fullname}
    \bibliography{main}
}

\clearpage
\appendix
\setcounter{page}{1}
\maketitlesupplementary 
% \clearpage
\setcounter{page}{1}
% \maketitlesupplementary

\begin{figure*}[t]
    \centering
    \includegraphics[width=\textwidth]{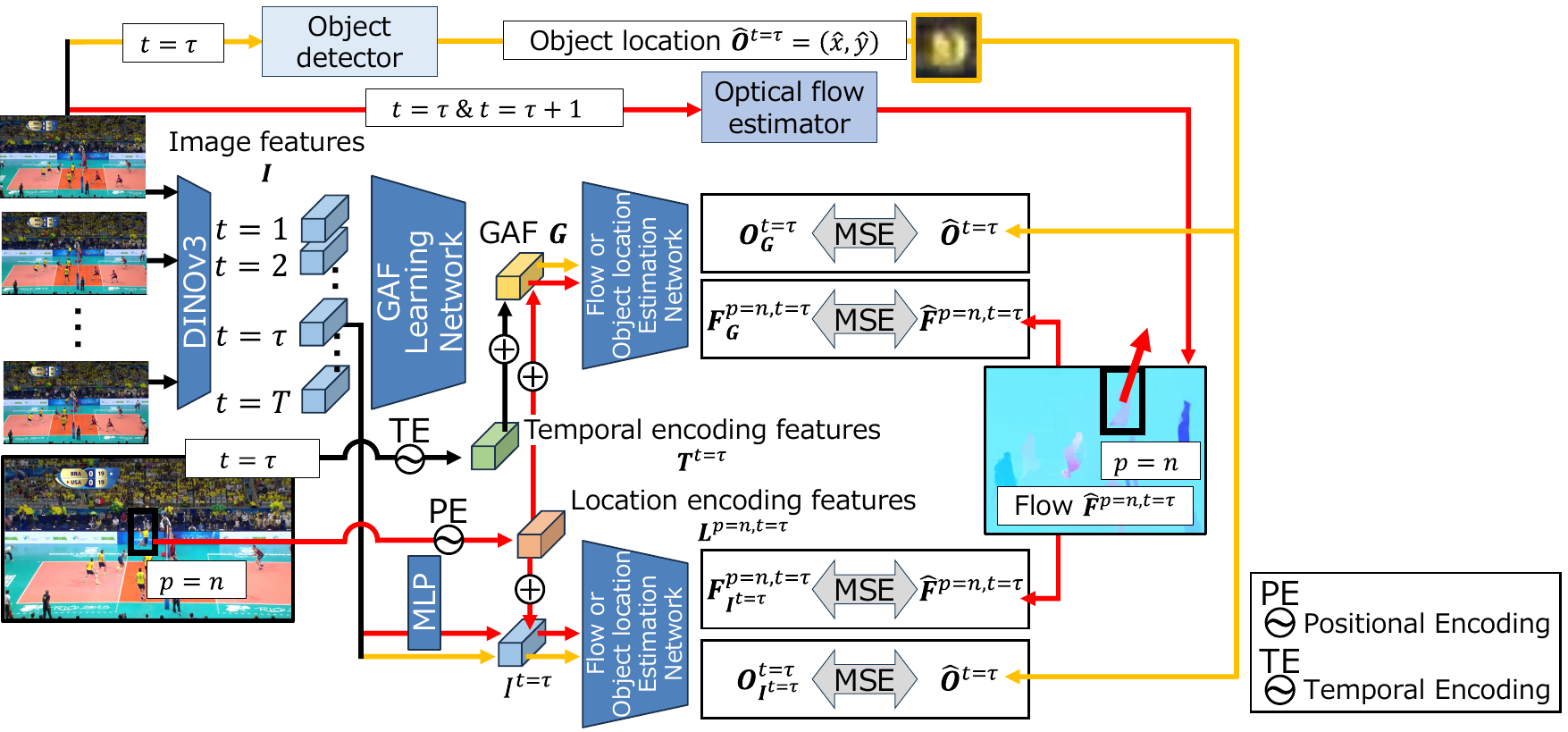}
     \caption{Detail of our pretext tasks. Our pretext tasks comprise person-flow estimation and group-relevant object location estimation. Person flow estimation utilizes a pre-trained optical flow estimator. Group-relevant object location estimation utilizes a pre-trained object detector. Since both pretext tasks use only these pre-trained models, training can be conducted in a self-supervised manner with pseudo labels. The black paths are shared by both pretext tasks, while the \textcolor{red}{\textbf{red}} paths are used only for person flow estimation, and the \textcolor{myyellow}{\textbf{yellow}} paths are used only for group-relevant object location estimation.}
     \label{fig:detail_loss}
\end{figure*}

\section{Implementation Details}
\label{sec:implement_detail}
This section presents implementation details not covered in the main paper.

\subsection{Pretext tasks}
\label{subsec:pretext}

Figure~\ref{fig:detail_loss} shows the details of our pretext tasks (i.e., person flow estimation and group-relevant object localization).
In person flow estimation, the optical flow image is obtained from two adjacent frames in a video (e.g., $\tau$ and $\tau+1$ in Fig.~\ref{fig:detail_loss}) with a pre-trained optical flow estimator (i.e., RAFT~[\textcolor{cvprblue}{44}] in our implementation). 
% Using the estimated optical flow, a pseudo label of flow values for each person is computed and used for $\mathcal{L}_{\mathrm{F}}$.
The pseudo labels of flow values for each person are obtained from the optical flow image by extracting the flow values of the center of the person's bounding box.
For group-relevant object localization, annotated and detected ball coordinates are used to compute $\mathcal{L}_{\mathrm{O}}$ for the Volleyball dataset (VBD) and NBA dataset (NBA), respectively.
In VBD and NBA, the $xy$ values of the center point of the bounding box are calculated from the bounding box coordinates as the pseudo labels (i.e., $\bm{\hat{O}}$) for $\mathcal{L}_{\mathrm{O}}$. 

\subsection{Experimental Conditions}
\label{subsec:ex_condi}

We trained our models on a single NVIDIA RTX A6000 GPU with a batch size of 8.
% 
% For the Volleyball dataset (VBD), we follow the experimental setting of GAFL~[\textcolor{cvprblue}{28}] and use video clips consisting of 10 frames (i.e., $T=10$). 
% 
% For the NBA dataset (NBA), we use 12 frames (i.e., $T=12$).
For VBD, we use video clips consisting of 10 frames (i.e., $T=10$) as with GAFL~[\textcolor{cvprblue}{28}].
For NBA, we use 12 frames (i.e., $T=12$).
% For the NBA dataset (NBA), we use 12 frames (i.e., $T=12$) due to GPU memory limitations, whereas Dual-AI~[\textcolor{cvprblue}{13}] uses 20 frames.
%
On each dataset, the same number of frames is used for all compared methods to ensure a fair comparison.
% We use the same frame-number setting across all compared methods for a fair comparison.
% 
For training the $\mathcal{L}_{\mathrm{F}}$, we use $N=12$ for VBD and $N=10$ for NBA, since 12 players are observed in most images in VBD and 10 in NBA. 
For inpainting, given the $xy$ coordinates of a ball, we define the mask as a circle centered at the corresponding location, where the radius is manually fixed for all experiments.
In our experiments with YOLOX~[\textcolor{cvprblue}{12}] to detect person bounding boxes, we use the publicly released Deep-EIoU~\cite{Deep-EIoU} weights.
The code is available at~\url{https://github.com/hsiangwei0903/Deep-EIoU}.
For the comparison of group activity retrieval, we use HRN~[\textcolor{cvprblue}{16}] and GAFL~[\textcolor{cvprblue}{28}]. These codes are available at~\url{https://github.com/chihina/GAFL-CVPR2024}.

\section{Additional Experiments}
\label{sec:add_ex}

This section presents additional experiments that could not be included in the main paper.

\begin{table}[t]
    \centering
    \caption{Comparison with state-of-the-art self-supervised GAF learning methods trained with supervision of person action classes on VBD. All other results are from~[\textcolor{cvprblue}{28}].}
    \begin{tabular}{l|ccc}
    \hline
                                       & \multicolumn{3}{c}{VBD}                       \\ \hline
    Method                             & Hit@1         & Hit@2         & Hit@3         \\ \hline
    HiGCIN~[\textcolor{cvprblue}{57}]  & 50.0          & 66.3          & 74.5          \\
    DIN~[\textcolor{cvprblue}{58}]     & 57.0          & 73.1          & 81.1          \\
    Dual-AI~[\textcolor{cvprblue}{13}] & 64.4          & 76.5          & 82.0          \\
    GAFL~[\textcolor{cvprblue}{28}]    & \textbf{84.8} & 89.6          & 91.8          \\ \hline
    Ours                               & 82.7          & \textbf{90.0} & \textbf{93.0} \\ \hline
    \end{tabular}
    \label{tab:comp_garet_pac_sup}
\end{table}
%%%%%%%%%%%%%%%%%%%%%%%%%%%%%%

\subsection{Evaluation in Group Activity Retrieval}
\label{subsection:ret_sup}

\subsubsection{Comparative experiments}
\label{sebsubsection:comparative_sup}

\if 0
Table~\ref{tab:comp_garet_sup} shows additional results (i.e., Hit@2) that could not be included in Table~\textcolor{cvprblue}{1} of the main paper due to page limitations. 
% 
As with the results in Hit@1 and Hit@3 shown in the main paper, we can see that our method outperforms other methods in Hit@2 on VBD and NBA.
\fi

While our work proposes two pretext tasks (i.e., person flow estimation and group-relevant object location estimation), another pretext task estimates person action classes, as proposed in~[\textcolor{cvprblue}{28}].
All of these three pretext tasks can be achieved with pretrained estimators (i.e., flow estimators for person flow estimation, object detectors for group-relevant object location estimation, and action recognizers for person action recognition).
Therefore, to demonstrate the effectiveness of our two pretext tasks compared with the one with person action classification, our method is compared with four networks that are trained with person action classification.

%Our method is compared with previous self-supervised GAF learning methods using person action classes, as GAFL~[\textcolor{cvprblue}{28}] proposes learning $\bm{G}$ by estimating person action classes. For this estimation of person action classes, GAFL uses supervision with annotated person action classes.
%Table~\ref{tab:comp_garet_pac_sup} shows the comparison of the previous pretext task using person action classes and our pretext tasks (i.e., person flow estimation and group-relevant object location estimation) on VBD.

Table~\ref{tab:comp_garet_pac_sup} shows the results on VBD.
While all methods except ``Ours'' train $\bm{G}$ with person-action supervision from manual annotations, ``Ours'' utilizes pseudo labels estimated by person flow and group-relevant object location estimation methods.
The results show that our method is better than the other three methods (i.e., HiGCIN~[\textcolor{cvprblue}{57}], DIN~[\textcolor{cvprblue}{58}], and Dual-AI~[\textcolor{cvprblue}{13}]) and achieves comparable performance to GAFL. 
% 
%These results validate that our pretext tasks are suitable for learning group activities compared with the pretext task with the person action classes.
% 
Even though these four previous methods are trained with the ground truth labels of person actions, our method achieves good performance only with estimated person flows and object locations, thereby eliminating the need for the manual annotations of action class labels.

\subsubsection{Ablation studies on Inpainting}
\label{subsubsection:effect_inpaint_sup}

\begin{table}[t]
    \centering
    \caption{Detailed analysis of inpainting on VBD and NBA. Unlike Table~\textcolor{cvprblue}{3} of the main paper, results when only $\mathcal{L}_{\mathrm{O}}$ is used for training are shown for further analysis.}
    \begin{tabular}{l|c|c}
    \hline
    \multirow{2}{*}{Input image} & VBD               & NBA                       \\ \cline{2-3}
    & Hit@1         & Hit@1         \\ \hline
    Original          & 72.9          & 30.5          \\ \hline
    Masking           & 45.0          & 18.5          \\ \hline\hline
    Inpainting (Ours) & \textbf{74.0} & \textbf{40.3} \\ \hline 
    \end{tabular}
    \label{tab:mask_comp_vbd_ball}
\end{table}

\if 0
\begin{table}[t]
    \centering
    \caption{
    Effects of inpainting during inference in our GAF learning on VBD and NBA. 
    }
    \begin{tabular}{ccc|c|c}
    \hline
    \multicolumn{2}{c}{Training}                                 & Inference             & VBD           & NBA           \\ \hline
    $\mathcal{L}_{\mathrm{F}}$ & $\mathcal{L}_{\mathrm{O}}$ &                       & Hit@1         & Hit@1         \\ \hline
    $\mathrm{\checkmark}$         & $\mathrm{\checkmark}$        & $\mathrm{\checkmark}$ & 82.4          & 44.8              \\
    $\mathrm{\checkmark}$         & $\mathrm{\checkmark}$        & -                     & 82.2          & 46.1          \\
    -                             & $\mathrm{\checkmark}$        & $\mathrm{\checkmark}$ & 81.5          & 43.4          \\
    -                             & $\mathrm{\checkmark}$        & -                     & 82.7 & 43.9 \\ \hline
    \end{tabular}
    \label{tab:infere_inpaint}
\end{table}
\fi

\paragraph{Effectiveness of Inpainting with only $\mathcal{L}_{\mathrm{O}}$.}
Table~\ref{tab:mask_comp_vbd_ball} shows the effectiveness of group-relevant object inpainting during training.
Different from the results obtained with $\mathcal{L}_{\mathrm{F}}$ and $\mathcal{L}_{\mathrm{O}}$ shown in Table~\textcolor{cvprblue}{3} of the main paper, Table~\ref{tab:mask_comp_vbd_ball} shows results with only $\mathcal{L}_{\mathrm{O}}$ on VBD and NBA for further analysis because this inpainting may affect only $\mathcal{L}_{\mathrm{O}}$.
As with Table~\textcolor{cvprblue}{3} of the main paper, in ``Original,'' original whole images are directly fed into the network to extract a GAF without inpainting of a ball. 
In ``Masking,'' the region of a ball in the whole image is filled with black pixels to remove the appearance of the ball.
In ``Inpainting (Ours),'' the region of a ball is masked via inpainting. 

Compared with ``Original,'' our method (i.e., ``Inpainting (Ours)'') is 1.1\% (1.1\%=74.0\%-72.9\%) better on VBD and 9.8\% (9.8\%=40.3\%-30.5\%) better on NBA. 
Additionally, we observe a significant performance improvement of our method compared to “Masking” on VBD and NBA. 
This is because the network can easily localize the ball from the appearance cues of the masked region in “Masking,” and this localization cannot contribute to our GAF learning.
These results clearly verify that ball inpainting is effective for our GAF learning with $\mathcal{L}_{\mathrm{O}}$.

\begin{table}[t]
    \centering
    \caption{
    Effects of inpainting during inference in our GAF learning on VBD and NBA. $\mathrm{\checkmark}$ indicates that the inpainting is applied during inference.
    }
    \begin{tabular}{c|c|c}
    \hline
    \multirow{2}{*}{\begin{tabular}[c]{@{}c@{}}Inpainting\\ during inference\end{tabular}} & VBD   & NBA   \\ \cline{2-3} 
                                                                                        & Hit@1 & Hit@1 \\ \hline
    $\mathrm{\checkmark}$                                                               & 81.5  & 43.4  \\
    - (Ours)                                                                                   & \textbf{82.7}  & \textbf{43.9}  \\ \hline
    \end{tabular}
    \label{tab:infere_inpaint}
\end{table}

\paragraph{Effectiveness of Inpainting during Inference.}
Table~\ref{tab:infere_inpaint} shows the results when we also apply group-relevant object inpainting at inference time. 
In our method, we do not apply inpainting during inference due to the additional computational cost incurred by the object detector and the inpainting model. 
However, since our method applies inpainting during training, it may introduce a domain gap when it is not applied during inference. 

Despite eliminating this domain gap, the results show that inpainting at inference degrades retrieval performance. 
This suggests that, in inference, the potential domain gap is less harmful than the errors introduced by imperfect ball localization and inpainting artifacts.

% \paragraph{Effect of location-guidance in our GAF learning}
\subsubsection{Ablation studies on location-guidance}
\label{subsubsection:effect_loc_guid_sup}

Table~\ref{tab:sup_loc} shows the ablation results on the location-guidance (i.e., $\bm{L}^{p,t}$ used in Eq.~\textcolor{cvprblue}{1},~\textcolor{cvprblue}{2}) in our method. 
Under both training settings, namely using only $\mathcal{L}_{\mathrm{F}}$ and jointly using $\mathcal{L}_{\mathrm{F}}$ and $\mathcal{L}_{\mathrm{O}}$, ``Ours'' outperforms ``Ours w/o $\bm{L}^{p,t}$'' on VBD and NBA. 
The significant performance gain suggests that ``Ours w/o $\bm{L}^{p,t}$'' only captures coarse motion over the whole image but fails to focus on fine-grained motions of each person in an image. 
We can interpret that location-guidance in our method explicitly encourages $\bm{G}$ to learn the fine-grained motions of each person.

\begin{table}[t]
    \centering
    \caption{
    Effects of the location-guidance in our GAF learning on VBD and NBA. Results obtained by the two settings (i.e., $\mathcal{L}_{\mathrm{F}}$ is only used or both $\mathcal{L}_{\mathrm{F}}$ and $\mathcal{L}_{\mathrm{O}}$ are used) are separated by the line.
    }
    \begin{tabular}{l|cc|c|c}
    \hline
                            & \multicolumn{2}{c|}{Loss}                                                    & VBD           & NBA           \\ \hline
    Method                  & \multicolumn{1}{c|}{$\mathcal{L}_{\mathrm{F}}$} & $\mathcal{L}_{\mathrm{O}}$ & Hit@1         & Hit@1         \\ \hline
    Ours w/o $\bm{L}^{p,t}$ & \multicolumn{1}{c|}{$\mathrm{\checkmark}$}      & -                          & 63.4          & 29.2          \\
    Ours                    & \multicolumn{1}{c|}{$\mathrm{\checkmark}$}      & -                          & \textbf{75.4} & \textbf{34.1} \\ \hline
    Ours w/o $\bm{L}^{p,t}$ & \multicolumn{1}{c|}{$\mathrm{\checkmark}$}      & $\mathrm{\checkmark}$      & 79.0          & 37.7          \\
    Ours                    & \multicolumn{1}{c|}{$\mathrm{\checkmark}$}      & $\mathrm{\checkmark}$      & \textbf{82.7} & \textbf{43.9} \\ \hline
    \end{tabular}
    \label{tab:sup_loc}
\end{table}

\begin{table}[t]
    \centering
    \caption{Additionl ablation studies on VBD and NBA.
    }
    % \vspace*{-3mm}
    \begin{tabular}{c|cc|cc|c|c}
    \hline
    \multicolumn{1}{l|}{} & \multicolumn{2}{c|}{$\mathcal{L}_{\mathrm{F}}$}                                              & \multicolumn{2}{c|}{$\mathcal{L}_{\mathrm{O}}$}                                              & VBD   & NBA   \\ \cline{2-7} 
                          & \multicolumn{1}{c|}{$\mathcal{L}_{\mathrm{F}, \bm{G}}$} & $\mathcal{L}_{\mathrm{F}, \bm{I}}$ & \multicolumn{1}{c|}{$\mathcal{L}_{\mathrm{O}, \bm{G}}$} & $\mathcal{L}_{\mathrm{O}, \bm{I}}$ & Hit@1 & Hit@1 \\ \hline
    (a)                   & \multicolumn{2}{c|}{-}                                                                       & \multicolumn{2}{c|}{-}                                                                       & 43.0  & 22.6  \\ \hline
    (j)                   & \multicolumn{1}{c|}{-}                                  & $\mathrm{\checkmark}$              & \multicolumn{2}{c|}{-}                                                                       & 71.3  & 29.5  \\ \hline
    (k)                   & \multicolumn{2}{c|}{-}                                                                       & \multicolumn{1}{c|}{-}                                  & $\mathrm{\checkmark}$              & 72.5  & 37.0  \\ \hline
    \end{tabular}
    \label{tab:abl_auxiliary}
    % \vspace*{-2mm}
\end{table}

\subsubsection{Effectiveness of Auxiliary Losses and Transformer}
\label{subsubsection:effect_auxi}
The Transformer encoder (shown in (b) of Fig.~\textcolor{cvprblue}{2} of the main paper) is trained from scratch but is not optimized with the auxiliary losses.
It is therefore impossible to train our network solely with them.
Instead, this Transformer is replaced by max pooling in (j) and (k) of Table~\ref{tab:abl_auxiliary}.
Even the auxiliary losses alone contribute to performance improvements, while the Transformer further improves performance, as shown in (b) and (c) of Table~\textcolor{cvprblue}{2} of the main paper.
%

\if 0
\begin{table}[t]
    \centering
    \caption{Comparison of GAFL with different image feature extractors. We replace VGG19, the original image feature extractor in GAFL, with DINOv3
    }
    \vspace*{-2mm}
    \scalebox{0.9}{
    % {\renewcommand{\arraystretch}{0.95}
    \begin{tabular}{lc|c|c}
    \hline
    \multicolumn{1}{l|}{Method}                                           & Backbone & Hit@1 on VBD        & Hit@1 on NBA         \\ \hline
    \multicolumn{1}{l|}{\multirow{2}{*}{GAFL~[\textcolor{cvprblue}{28}]}} & VGG19    & 61.1          & 24.7          \\
    \multicolumn{1}{l|}{}                                                 & DINOv3   & 38.1          & 25.3             \\ \hline
    \multicolumn{1}{l|}{Ours}                                             & DINOv3   & \textbf{82.7} & \textbf{43.9} \\ \hline
    \end{tabular}
    % }
    }
    \label{tab:gafl_dino}
    % \vspace*{-2mm}
\end{table}
\fi

\begin{table}[t]
    \centering
    \caption{Comparison of GAFL with different image feature extractors. We replace VGG19, the original image feature extractor in GAFL, with DINOv3
    }
    % \vspace*{-2mm}
    % \scalebox{0.9}{
    \begin{tabular}{l|c|c|c}
    \hline
    \multirow{2}{*}{Method}                                                      & \multicolumn{1}{l|}{\multirow{2}{*}{Backbone}} & VBD           & NBA           \\ \cline{3-4} 
                                                                                 & \multicolumn{1}{l|}{}                          & Hit@1         & Hit@1         \\ \hline
    \multirow{2}{*}{GAFL~[\textcolor{cvprblue}{28}]} & VGG19                                          & 61.1          & 24.7          \\
                                                                                 & DINOv3                                         & 38.1          & 25.3          \\ \hline
    Ours                                                                         & DINOv3                                         & \textbf{82.7} & \textbf{43.9} \\ \hline
    \end{tabular}
    % }
    \label{tab:gafl_dino}
    % \vspace*{-2mm}
\end{table}

\begin{table}[t]
    \centering
    \caption{
    Effects of architecture size for GAF learning on VBD and NBA. Results obtained by GAFL~[\textcolor{cvprblue}{28}] and our method are separated by a double line. In our method, results obtained by different ViTs in DINOv3 are shown for further discussion.
    }
    \scalebox{0.8}{
    {\renewcommand{\arraystretch}{0.95}
    \begin{tabular}{l|cc|cc|cc}
    \hline
    \multirow{2}{*}{Method}         & \multirow{2}{*}{\begin{tabular}[c]{@{}c@{}}All\\ Params\end{tabular}} & \multirow{2}{*}{\begin{tabular}[c]{@{}c@{}}Learnable\\ Params\end{tabular}} & \multicolumn{2}{c|}{Time[ms]} & \multicolumn{2}{c}{Hit@1}                          \\ \cline{4-7} 
                                    &                                                                       &                                                                             & \multicolumn{1}{c|}{VBD}    & NBA    & \multicolumn{1}{c|}{VBD}           & NBA           \\ \hline
    GAFL~[\textcolor{cvprblue}{28}] & 79M                                                                   & 68M                                                                         & \multicolumn{1}{c|}{82.4}   & 86.3   & \multicolumn{1}{c|}{61.1}          & 24.7          \\ \hline\hline
    Ours (ViT-B)                    & 99M                                                                   & 21M                                                                         & \multicolumn{1}{c|}{86.6}   & 102.9  & \multicolumn{1}{c|}{73.8}          & 38.8          \\
    Ours (ViT-L)                    & 326M                                                                  & 48M                                                                         & \multicolumn{1}{c|}{272.3}  & 327.7  & \multicolumn{1}{c|}{\textbf{82.7}} & \textbf{43.9} \\ \hline
    \end{tabular}
    }
    }
    \label{tab:arch_size}
\end{table}

\subsubsection{Effect of Architectures and Training Strategies}
\label{subsubsection:arc_and_train}

\paragraph{Effect of Backbone Choice.}
% As shown in Table~\ref{tab:gafl_dino}, GAFL performs worse with DINOv3. This may be because it is not easy to fully exploit the potential of DINO by merely swapping another feature extractor for DINO, as verified in~[\textcolor{cvprblue}{67}].
%
Table~\ref{tab:gafl_dino} shows the results when replacing VGG19, the original image feature extractor in GAFL~[\textcolor{cvprblue}{28}], with DINOv3. As shown in the table, GAFL performs worse with DINOv3. This suggests that merely replacing the feature extractor with DINOv3 is not sufficient to fully exploit its potential, as also observed in~[\textcolor{cvprblue}{67}].

\paragraph{Comparison of DINOv3 architecture.}

\if 0
Table~\ref{tab:arch_size} shows the architecture sizes of DINOv3 and the results when replacing the feature extractor with ViT-B and ViT-L. For ViT-B, we train the last block of its ViT.
%
While ViT-L yields stronger results, we confirm that ViT-B is still sufficient to achieve state-of-the-art performance. Moreover, our design effectively exploits the capability of large-scale pretrained feature extractors, achieving high performance while keeping the number of learnable parameters smaller than GAFL.
%
This favorable trade-off between performance and efficiency is also supported by the inference-time comparison.
Table~\ref{tab:inference_time} shows that Ours (ViT-B) outperforms GAFL~[\textcolor{cvprblue}{28}] with a similar runtime. 
\fi

Table~\ref{tab:arch_size} shows the architecture sizes of DINOv3, the results when replacing the feature extractor with ViT-B and ViT-L, and the inference-time comparison with GAFL~[\textcolor{cvprblue}{28}]. For ViT-B, we train the last block of its ViT. 
While ViT-L yields stronger results, we confirm that ViT-B is still sufficient to achieve state-of-the-art performance. Moreover, our design effectively exploits the capability of large-scale pretrained feature extractors, achieving high performance while keeping the number of learnable parameters smaller than GAFL.
In addition, Ours (ViT-B) outperforms GAFL~[\textcolor{cvprblue}{28}] with maintaining a similar runtime.

\begin{table}[t]
    \centering
    \caption{
    Comparison of training strategies for DINOv3 in our method. Results obtained by our method are shown in (i). In (a), (b), (c), and (i), we fine-tuned different numbers of transformer blocks in the ViT of DINOv3. In (d), DINOv3 is updated by LoRA~\cite{LoRA}. In (e), (f), (g), and (h), an additional linear layer or transformer blocks are attached to the frozen DINOv3 for fine-tuning. 
    }
    \scalebox{0.75}{
    \begin{tabular}{c|cccc|c|c}
    \hline
    \multicolumn{1}{l|}{} & \multicolumn{4}{c|}{Training setting}                                                                                                                                                                                              & VBD           & NBA           \\ \hline
                          & \multicolumn{1}{c|}{\begin{tabular}[c]{@{}c@{}}Unfrozen\\ Blocks\end{tabular}} & \multicolumn{1}{c|}{LoRA}                  & \multicolumn{1}{c|}{Adapter}         & \begin{tabular}[c]{@{}c@{}}Learnable\\ Params\end{tabular} & Hit@1         & Hit@1         \\ \hline
    (a)                   & \multicolumn{1}{c|}{-}                                                         & \multicolumn{1}{c|}{-}                     & \multicolumn{1}{c|}{-}               & 23M                                                        & 64.6          & 30.0          \\
    (b)                   & \multicolumn{1}{c|}{24}                                                        & \multicolumn{1}{c|}{-}                     & \multicolumn{1}{c|}{-}               & 36M                                                        & 78.2          & 40.3          \\
    (c)                   & \multicolumn{1}{c|}{22, 23, 24}                                                & \multicolumn{1}{c|}{-}                     & \multicolumn{1}{c|}{-}               & 61M                                                        & 80.9          & 36.5          \\
    (d)                   & \multicolumn{1}{c|}{-}                                                         & \multicolumn{1}{c|}{$\mathrm{\checkmark}$} & \multicolumn{1}{c|}{-}               & 24M                                                        & 76.6          & 42.3          \\
    (e)                   & \multicolumn{1}{c|}{-}                                                         & \multicolumn{1}{c|}{-}                     & \multicolumn{1}{c|}{1 linear}        & 24M                                                        & 65.7          & 34.4            \\
    (f)                   & \multicolumn{1}{c|}{-}                                                         & \multicolumn{1}{c|}{-}                     & \multicolumn{1}{c|}{1 trans. block}  & 36M                                                        & 68.3          & 38.2            \\
    (g)                   & \multicolumn{1}{c|}{-}                                                         & \multicolumn{1}{c|}{-}                     & \multicolumn{1}{c|}{2 trans. blocks} & 48M                                                        & 70.2          & 41.0            \\
    (h)                   & \multicolumn{1}{c|}{-}                                                         & \multicolumn{1}{c|}{-}                     & \multicolumn{1}{c|}{3 trans. blocks} & 61M                                                        & 70.3          & 41.4            \\ \hline
    (i)                   & \multicolumn{1}{c|}{23, 24}                                                    & \multicolumn{1}{c|}{-}                     & \multicolumn{1}{c|}{-}               & 48M                                                        & \textbf{82.7} & \textbf{43.9} \\ \hline
    \end{tabular}
    }
    \label{tab:sup_how_to_learn}
\end{table}

\begin{table}[t]
    \centering
    \caption{Comparison of input features on VBD and NBA. While ``Cropped'' utilizes cropped-image features as input, our method utilizes whole-image features for GAF learning.}
    \begin{tabular}{l|c|c}
    \hline
    \multirow{2}{*}{Input features} & VBD           & NBA           \\ \cline{2-3} 
                                    & Hit@1         & Hit@1         \\ \hline
    Cropped                         & 58.8          & 28.8          \\ \hline
    Whole (Ours)                    & \textbf{74.0} & \textbf{40.3} \\ \hline
    \end{tabular}
    \label{tab:crop_garet}
\end{table}

\paragraph{Comparison of training strategies.}
Table~\ref{tab:sup_how_to_learn} shows the results with various training strategies of DINOv3 in our method. 
In this table, the results obtained by our method are shown in (i).
% 
% To examine which training strategy is suitable for learning group activity feat with DINOv3 (ViT-L), we varied the set of applied Low Rank Adaptation (LoRA)~\cite{DBLP:conf/iclr/LoRA} and learnable adapters. 
% We varied the Low Rank Adaptation (LoRA)~\cite{DBLP:conf/iclr/LoRA} and learnable adapters. 
%
% We first investigate how fine-tuning DINOv3 itself affects performance. ViT-L consists of 24 transformer blocks, and Figs. (a), (b), (c), and (i) show the results when the ViT is kept frozen, or when only the last one, two, or three blocks are made learnable, respectively. Among these settings, (i) achieves the best performance. This indicates that fine-tuning blocks of DINOv3 is beneficial for learning group activity features, and that updating only the last two blocks is the most effective setting. This may be because it adapts DINOv3 to the task while keeping most of its pretrained knowledge.
In (a), (b), (c), and (i), we compared the retrieval performance with different numbers of unfrozen blocks within 24 transformer blocks of DINOv3 (ViT-L).
% 
% Among these settings, (i) achieves the best performance. This indicates that fine-tuning blocks of DINOv3 is beneficial for learning group activity features, and that updating only the last two blocks is the most effective setting. This may be because it adapts DINOv3 to the task while keeping most of its pretrained knowledge.
Among these settings, (i) achieves the best performance. These results indicate that fine-tuning DINOv3 by updating only the last two transformer blocks is effective for keeping the pre-trained knowledge.

% We also examine the effectiveness of fine-tuning DINOv3 with LoRA in (d). LoRA updates the backbone while keeping the number of additional learnable parameters small by learning weight update matrices as low rank decompositions. In the DINOv3 backbone, we updated the query and value projections with rank 16. 
% When fine-tuning with LoRA, we can substantially improve performance with only a small number of additional learnable parameters, although it still does not surpass setting (i). This suggests that fine-tuning with LoRA is particularly useful in scenarios where the number of learnable parameters is constrained.
We also examine the fine-tuning of DINOv3 with Low Rank Adaptation (LoRA)~\cite{LoRA} in (d).
We update the query and value projections with rank 16 by LoRA in (d).
Compared with (d), (i) is still better than (d). The results demonstrate that fine-tuning only the last two transformer blocks is better than updating a small adapter in (d). 

% We next investigate whether it is effective to keep DINOv3 frozen and instead learn additional adapters in (e), (f), (g), and (h). ``1 linear'' feeds the CLS token into a one-linear layer or ``n trans. block(s)'' passes the DINOv3 output through additional $n$ transformer layers, following the setup in~\cite{DBLP:conf/cvpr/gaze-lle, DBLP:conf/cvpr/dino_txt}. 
Finally, we conduct fine-tuning with additional adapters while keeping DINOv3 frozen in (e), (f), (g), and (h). As in~\cite{Dinotext, gazelle}, ``1 linear'' feeds the CLS token into one linear layer or ``n trans. block(s)'' passes the CLS token and patch tokens into additional $n$ transformer blocks.
% 
% In (e), we learn an additional linear layer, but the performance improvement is limited. For a comparable number of learnable parameters, applying LoRA as in (d) is more effective. In (f), (g), and (h), we train additional transformer blocks on top of the frozen backbone, yet the performance gains are modest relative to the increase in learnable parameters.
Compared with (i), (e) is worse than (i). Furthermore, the performance in (e) is also worse than (d), in which LoRA updates the same number of parameters.
In (f), (g), and (h), we added additional transformer blocks on the frozen backbone. The performance is better than (e) thanks to the large number of learnable parameters. However, (i) is still the best.
%
% Considering performance, the number of learnable parameters, and the total number of network parameters, we adopt a setting that fine-tunes only the last two blocks of ViT as our default setting.

Based on the discussion, our training strategy (i.e., fine-tuning the last two transformer blocks of ViT in DINOv3) is better than the other strategies. 
These results may come from the domain gap between the pre-training of DINO and downstream tasks.
In~\cite{Dinotext,gazelle}, DINO is fine-tuned for segmentation and gaze target detection in general scenes. Since such general scenes are used for pre-training of DINOv3, fine-tuning DINOv3 with an additional adapter is enough while freezing DINOv3. Unlike~\cite{Dinotext,gazelle}, we utilize sports images that are largely underrepresented in the pre-training of DINOv3. This domain gap suggests that directly fine-tuning DINOv3, as in~[\textcolor{cvprblue}{20}], is more effective than the other strategies.
% These results may come from the complexity of group activity. Fine-tuning of DINO with an adapter is effective for visual cue-based tasks (e.g., segmentation and gaze target detection in~\cite{Dinotext, gazelle}) because DINO is pre-trained to obtain fine-grained visual features. However, the pre-training is not enough for the high-level group activity understanding.

\subsubsection{Detailed analysis}
\label{subsubsection:det_ana}

\paragraph{Whole-image features vs. cropped-image features.}
Unlike previous self-supervised GAF learning methods~[\textcolor{cvprblue}{16, 28}], which take person bounding boxes cropped from an image as input, our method feeds the whole image into DINOv3 to capture global context learned in $\mathcal{L}_{\mathrm{O}}$.
To verify the effectiveness of global context learning by $\mathcal{L}_{\mathrm{O}}$, we conduct an experiment where only $\mathcal{L}_{\mathrm{O}}$ is used for GAF learning with cropped person bounding boxes, as shown in Table~\ref{tab:crop_garet}.
% We analyze the effect of the global context obtained from the whole image in Table~\ref{tab:crop_garet}. 
% Table~\ref{tab:crop_garet} shows the performance change when only cropped person bounding boxes are used without the whole image. 
%
% While existing methods often extract per-person features by applying RoIAlign~\cite{DBLP:conf/iccv/RoIAlign} to high-resolution CNN feature maps, our method employs a ViT as the image feature extractor, which produces a sequence of patch tokens rather than a dense convolutional feature map. 
% 
% Since RoIAlign is not directly applicable to such token sequences, we obtain the feature of each person by averaging the patch tokens that fall inside the bounding box of each person.
RoIAlign~\cite{RoIAlign} used in GAFL~[\textcolor{cvprblue}{28}] is not applied to DINOv3 due to the low-resolution patch tokens.
Therefore, we average the patch tokens within each person's bounding box to obtain cropped image features as with~\cite{goal, region-based}.
%
% This token-level pooling over patch embeddings is a common design choice in recent ViT-based region modeling and detection methods, where region features are computed by average-pooling patch tokens inside a bounding box~\cite{goal, DBLP:conf/cvpr/Shlapentokh-Rothman24}, and is therefore consistent with established practice rather than an ad-hoc operation.

% Table~\ref{tab:crop_garet} shows the results when we only utilize $\mathcal{L}_{\mathrm{O}}$ for the loss function. 
% Table~\ref{tab:crop_garet} shows the results when we only utilize $\mathcal{L}_{\mathrm{O}}$ for the loss function. 
% 
% In our method with whole image input, group-relevant objects are inpainted. 
%
In Table~\ref{tab:crop_garet}, our whole-image variant is better than ``Cropped'' on both VBD and NBA.
These results demonstrate that removing the appearance cues of the group-relevant objects by cropping person bounding boxes discards the global context. 

\begin{table}[t]
    \centering
    \caption{
    Effects of whole image flow estimation on VBD and NBA. While the optical flow image obtained from a whole image is estimated as a pretext task in ``Whole image flow,'' our method proposes to estimate flow values of each person as a pretext task for learning people-relevant features in $\bm{G}$.  
    }
    \begin{tabular}{l|cc|c|c}
    \hline
                       & \multicolumn{2}{c|}{Loss}                                                    & VBD           & NBA           \\ \hline
    Method             & \multicolumn{1}{c|}{$\mathcal{L}_{\mathrm{F}}$} & $\mathcal{L}_{\mathrm{O}}$ & Hit@1         & Hit@1         \\ \hline
    Whole image flow   & \multicolumn{1}{c|}{$\mathrm{\checkmark}$}      & -                          & 51.9          & 23.8          \\
    Person flow (Ours) & \multicolumn{1}{c|}{$\mathrm{\checkmark}$}      & -                          & \textbf{75.4} & \textbf{34.1} \\ \hline
    Whole image flow   & \multicolumn{1}{c|}{$\mathrm{\checkmark}$}      & $\mathrm{\checkmark}$      & 76.7          & 38.2          \\
    Person flow (Ours) & \multicolumn{1}{c|}{$\mathrm{\checkmark}$}      & $\mathrm{\checkmark}$      & \textbf{82.7} & \textbf{43.9} \\ \hline
    \end{tabular}
    \label{tab:sup_whole_flow}
\end{table}

\begin{table}[t]
    \centering
    \caption{
    Effects of errors in bounding boxes detected by YOLOX~[\textcolor{cvprblue}{12}] on VBD. ``GT'' indicates that the experiments utilize the ground-truth person bounding boxes. ``YOLOX'' indicates that the experiments utilize detected tracklets by Deep-EIoU~\cite{Deep-EIoU}.
    }
    \begin{tabular}{l|c}
    \hline
                                     & VBD   \\ \hline
    Method                                  & Hit@1 \\ \hline
    GAFL~[\textcolor{cvprblue}{28}] (YOLOX) & 51.5  \\
    GAFL~[\textcolor{cvprblue}{28}] (GT)    & 61.1  \\ \hline
    Ours (YOLOX)                            & 81.2  \\
    Ours (GT)                               & 82.7  \\ \hline
    \end{tabular}
    \label{tab:gt_yolo}
\end{table}

\paragraph{Effects of whole image flow estimation.}
In our method, flow estimation is applied for each person in an image to capture local dynamics. As a variant, our method can easily extend to estimate the optical flow of a whole image. To estimate optical flow over the whole image, the patch tokens obtained from DINOv3 are processed through a two-layer $1\times1$ convolutional network.

Table~\ref{tab:sup_whole_flow} shows the comparison between estimating the optical flow over a whole image and person flow. Our method outperforms ``Whole image flow'' in all metrics. These results validate that the optical flow estimation over the whole image is not suitable for GAF learning. 
This is because the flow over the whole image is dominated by background motion, whereas person flows are more important for representing group activities.

\begin{figure}[t]
    \centering
    \includegraphics[width=\columnwidth]{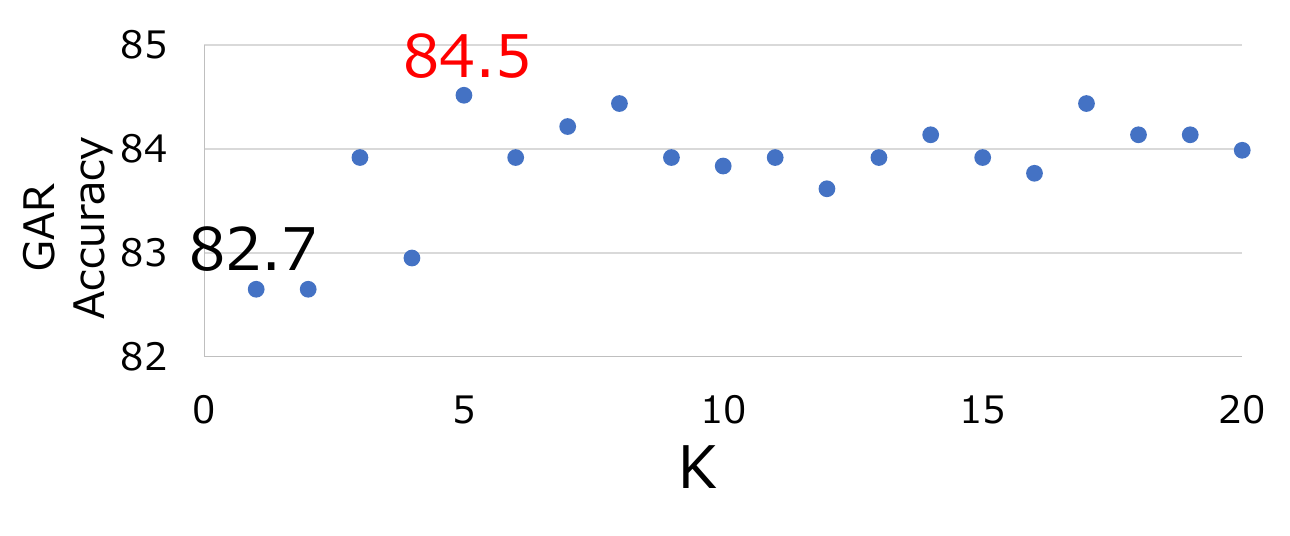}\\
    % \vspace*{-3mm}
    (a) Volleyball dataset (VBD)\\
    \includegraphics[width=\columnwidth]{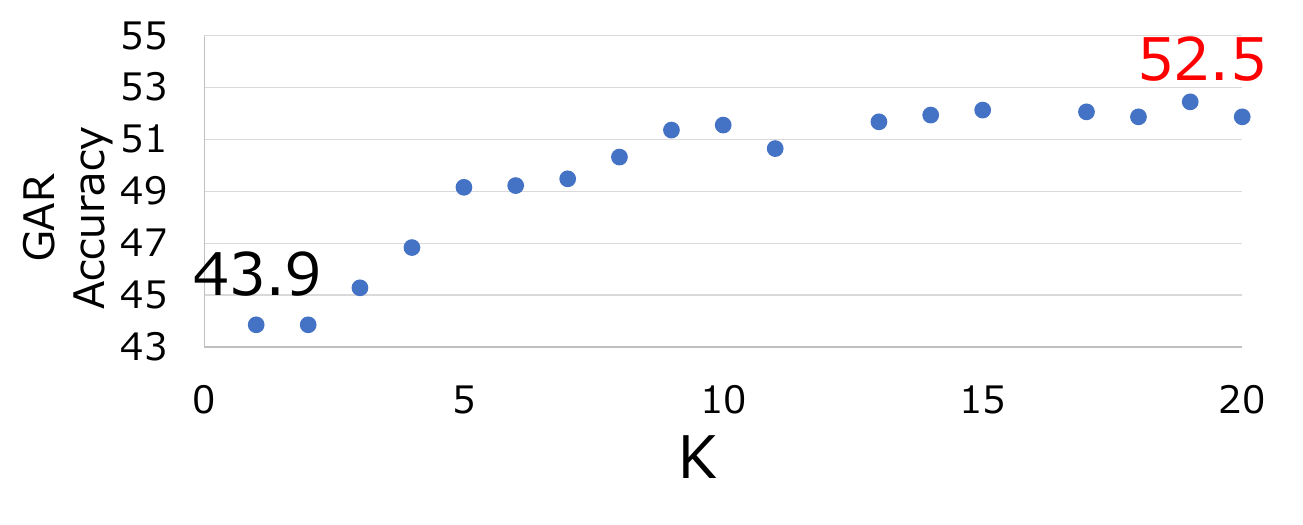}\\
    % \vspace*{-3mm}
    (b) NBA dataset (NBA)\\
    % \vspace*{-1mm}
    \caption{GAR accuracy curve by the KNN classification on VBD and NBA. K changes from 1 to 20 in our experiment.}
    \label{fig:KNN}
    % \vspace*{-2mm}
\end{figure}

\paragraph{Effects of errors in person detection.}

\if 0
\begin{table}[t]
    \centering
    \caption{
    Effects of errors in bounding boxes detected by YOLOX~[\textcolor{cvprblue}{12}]. ``GT'' indicates that the experiments utilize the ground-truth person bounding boxes. ``YOLOX'' indicates that the experiments utilize detected tracklets obtained by a people tracker (i.e., Deep-EIoU~\cite{Deep-EIoU}).
    }
    \begin{tabular}{l|ccc}
    \hline
    Dataset     & \multicolumn{3}{c}{VBD}                                       \\ \hline
    Method      & Hit@1 & \multicolumn{1}{l}{Hit@2} & \multicolumn{1}{l}{Hit@3} \\ \hline
    GAFL~[\textcolor{cvprblue}{28}] (YOLOX)    & 51.5          & 68.6 & 75.8         \\
    GAFL~[\textcolor{cvprblue}{28}] (GT)    & 61.1          & 75.1 & 82.4          \\ \hline\hline
    Ours (YOLOX)   & 81.2  & 89.5                      & 92.7                      \\
    Ours (GT) & 82.7  & 90.0                      & 93.0                      \\ \hline
    \end{tabular}
    \label{tab:gt_yolo}
\end{table}
\fi

To reveal the effects of errors in person detection, we utilize the detected person bounding boxes on VBD, instead of their GTs, as shown in Table~\ref{tab:gt_yolo}. 
The results show that our method achieves consistently high performance even with detected person bounding boxes by YOLOX~[\textcolor{cvprblue}{12}]. 
In ``GAFL,'' the retrieval performance is significantly dropped by the errors in the detected person bounding boxes.
From these results, we conclude that our method is more robust to detection errors in person bounding boxes compared with GAFL~[\textcolor{cvprblue}{28}].

\begin{figure}[t]
    % \vspace*{-2mm}
    \centering
    \includegraphics[width=\columnwidth]{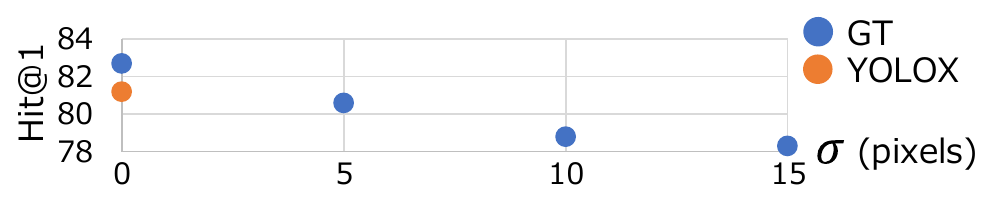}
    \vspace*{-7mm}
    \caption{Performance under different noise levels.}
    \label{fig:add_noise}
    \vspace*{-2mm}
\end{figure}

\paragraph{Robustness to noisy pseudo-labels.}
Figure~\ref{fig:add_noise} shows Hit@1 under different zero-mean Gaussian noise levels added to GT person bounding boxes and GT ball locations.
%in our method. 
% For reference, we also show Hit@1 when using person bounding boxes detected by YOLOX, which is also reported in Table~\textcolor{cvprblue}{14} of the Supp.
%($\sigma$ = $x$ pixels) 
Similar Hit@1 scores under $\sigma=10$ and $\sigma=15$ validate the robustness of our method to erroneous bounding boxes. We will also show a sensitivity analysis with respect to other pseudo-labels.

% small ver.
\if 0
\begin{table}[t]
    \centering
    \caption{Additional pretext goal-detection task on NBA.
    }
    \vspace*{-3mm}
    \scalebox{0.9}{
    {\renewcommand{\arraystretch}{0.9}
    \begin{tabular}{c|c|c|c}
    \hline
    $\mathcal{L}_{\mathrm{F}}$                & $\mathcal{L}_{\mathrm{O}}$ (Ball)                  & $\mathcal{L}_{\mathrm{O}}$ (Goal)                  & Hit@1 \\ \hline
%    -                          & \multicolumn{1}{c|}{$\mathrm{\checkmark}$} & -                     & 40.3  \\
%    -                          & \multicolumn{1}{c|}{$\mathrm{\checkmark}$} & $\mathrm{\checkmark}$ & 40.8  \\ \hline
    $\mathrm{\checkmark}$      & \multicolumn{1}{c|}{$\mathrm{\checkmark}$} & -                     & 43.9  \\
    $\mathrm{\checkmark}$      & \multicolumn{1}{c|}{$\mathrm{\checkmark}$} & $\mathrm{\checkmark}$ & 45.5  \\ \hline
    \end{tabular}
    }
    }
    \label{tab:add_goal}
    \vspace*{-2mm}
\end{table}
\fi

\begin{table}[t]
    \centering
    \caption{Additional pretext goal-detection task.
    }
    \vspace*{-2mm}
    \scalebox{0.9}{
    % {\renewcommand{\arraystretch}{0.9}
    \begin{tabular}{c|cc|c}
    \hline
    $\mathcal{L}_{\mathrm{F}}$ & \multicolumn{2}{c|}{$\mathcal{L}_{\mathrm{O}}$}                    & NBA   \\ \hline
    Person Flow                & \multicolumn{1}{c|}{Ball}                  & Goal                  & Hit@1 \\ \hline
   -                          & \multicolumn{1}{c|}{$\mathrm{\checkmark}$} & -                     & 40.3  \\
   -                          & \multicolumn{1}{c|}{$\mathrm{\checkmark}$} & $\mathrm{\checkmark}$ & \textbf{40.8}  \\ \hline
    $\mathrm{\checkmark}$      & \multicolumn{1}{c|}{$\mathrm{\checkmark}$} & -                     & 43.9  \\
    $\mathrm{\checkmark}$      & \multicolumn{1}{c|}{$\mathrm{\checkmark}$} & $\mathrm{\checkmark}$ & \textbf{45.5}  \\ \hline
    \end{tabular}
    }
    % }
    \label{tab:add_goal}
    \vspace*{-2mm}
\end{table}

%%%[\R2, \R3] \underline{\textbf{Limited generalization of the pretext ball-detection task.}
\paragraph{More localization tasks.}
%While only a ball is used as an activity-related object, a basketball goal system is also employed on NBA, as shown in Table~\ref{tab:add_goal}. $\mathcal{L}_{\text O}$ with the goal system further improves performance. We also plan to present experiments where a net system on VBD, court lines, and referees are used in $\mathcal{L}_{\text O}$.
While our method currently uses only a ball as an activity-related object, we also incorporate a basketball goal system on NBA (Table~\ref{tab:add_goal}). $\mathcal{L}_{\text O}$ with the goal system further improves performance. We also plan to include experiments using a net system on VBD, court lines, and referees in $\mathcal{L}_{\text O}$.

\begin{table}[t]
    \centering
    \caption{
    Comparison with supervised group activity recognition methods on VBD and NBA. The results of all methods are copied from their papers. The four methods above the middle line take the whole image as input, while the four methods between the middle line and ours use person bounding boxes together with the whole image in both training and inference. For a fair comparison, in all methods, the only manually annotated labels are group activity classes, and only images are used in inference.
    }
    \scalebox{0.93}{
    \begin{tabular}{lc|c|c}
    \hline
    \multirow{2}{*}{Method}                                            & \multicolumn{1}{l|}{\multirow{2}{*}{Extractor}} & VBD           & NBA           \\ \cline{3-4} 
                                                                       & \multicolumn{1}{l|}{}                           & Merged MCA    & MPCA          \\ \hline
    DFWSGAR~[\textcolor{cvprblue}{21}]     & ResNet-18                                       & 94.4          & 71.2          \\
    SOGAR~[\textcolor{cvprblue}{6]}        & ResNet-18                                       & 95.9          & 73.5          \\
    Flaming-Net~[\textcolor{cvprblue}{29]} & Inception-v3                                    & 95.2          & \textbf{76.0} \\
    LiGAR~[\textcolor{cvprblue}{4]}        & ResNet-18                                       & 76.1          & 57.1          \\ \hline
    SAM~[\textcolor{cvprblue}{56]}         & ResNet-18                                       & 93.1          & 51.5          \\
    Dual-AI~[\textcolor{cvprblue}{13]}     & Inception-v3                                    & 95.8          & 50.2          \\
    KRGFormer~[\textcolor{cvprblue}{32]}   & Inception-v3                                    & 95.0          & 67.1          \\
    MP-GCN~[\textcolor{cvprblue}{26]}      & YOLOV8x                                         & \textbf{96.1} & 74.6          \\ \hline\hline
    Ours                                                               & DINOv3                                          & \textbf{96.1} & 68.2          \\ \hline
    \end{tabular}
    }
    \label{tab:comp_wsgar_nba}
\end{table}

\begin{figure*}[t]
    \centering
    \includegraphics[width=\textwidth]{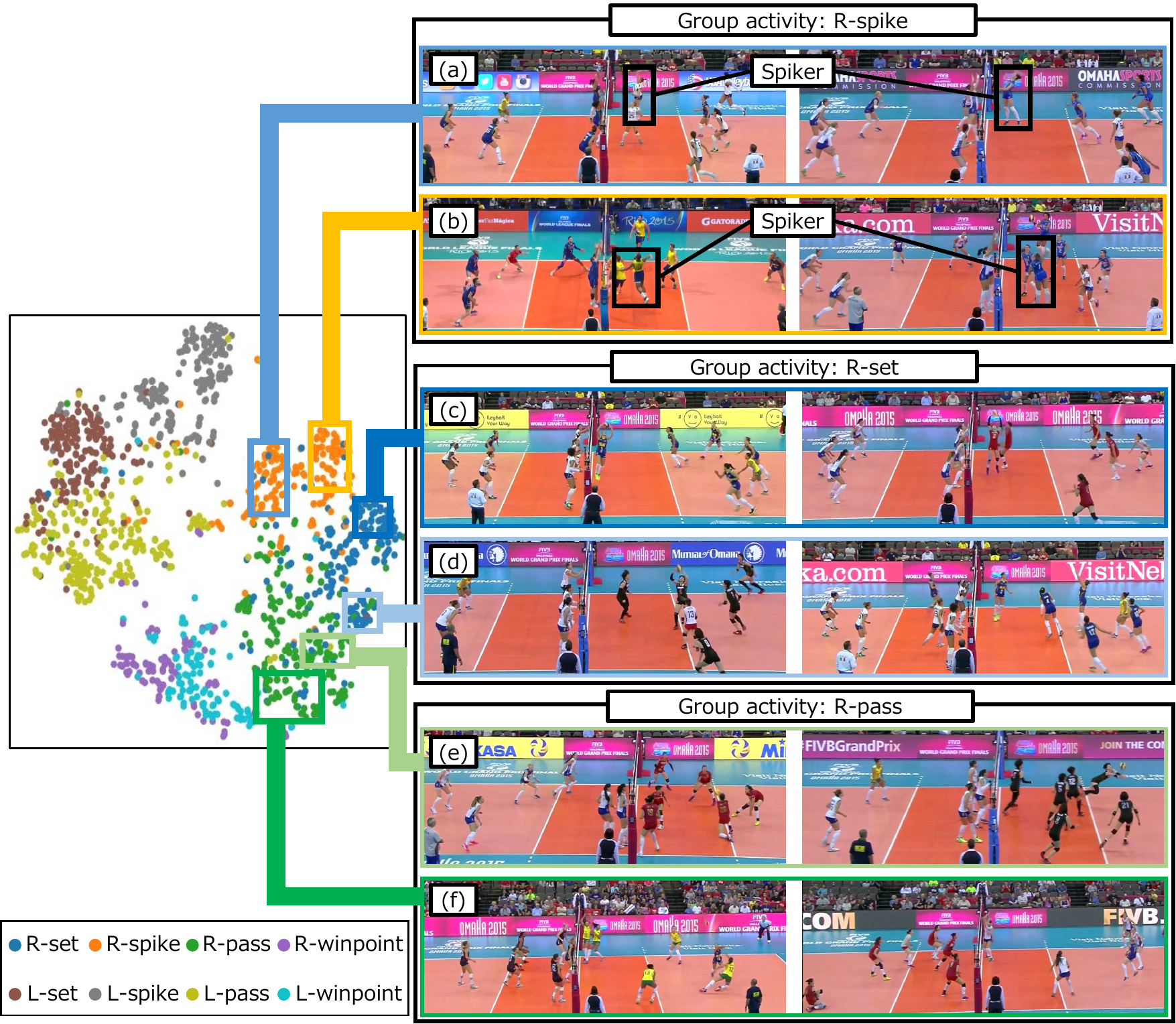}
     \caption{
     Visualization of the learned GAF space on VBD. 
     Unlike Fig.~\textcolor{cvprblue}{10} of the main paper, we also show actual images corresponding to the GAFs. The orange data points (i.e., ``R-spike'') are divided into (a) and (b) based on the spiker's position. The blue data points (i.e., ``R-set'') are divided into (c) and (d) based on the setter's position as similar to (a) and (b). In (e) and (f), the green data points (i.e., ``R-pass'') are split into two clusters that differ in terms of the receiver's position.
     }
     \label{fig:vis_fine}
\end{figure*}

\subsection{Evaluation in Group Activity Recognition}
\label{subsection:reco_sup}

\subsubsection{KNN for Group Activity Recognition}
\label{subsubsection:knn}
% \paragraph{KNN for Group Activity Recognition}
%
In addition to Group Activity Recognition (GAR) via the 1-nearest neighbor retrieval results shown in Fig.~\textcolor{cvprblue}{7} of the main paper, Fig.~\ref{fig:KNN} shows GAR accuracy for different numbers of neighbors K in KNN classification. Following~[\textcolor{cvprblue}{28}], we vary $K$ from 1 to 20. 

For VBD, $K>3$ achieves higher GAR accuracy than $K=1$, and the best result with $K=5$ is $1.8\%$ better than $K=1$. As with the results above, for NBA, $K>3$ outperforms $K=1$, and the best result with $K=19$ improves accuracy by $8.6\%$. 
These results demonstrate that using KNN for GAR is a simple yet effective approach, and that using a larger K is beneficial when the 1-NN accuracy is relatively low, as in NBA with $43.9\%$, compared with the already high accuracy on VBD with $82.7\%$.

% We attribute this difference to dataset characteristics. 
% 
% NBA sequences are longer than those in VBD, so each GAF includes longer temporal dynamics.
%  
% As a consequence, the encoded GAF includes common features related to people's behavior occurring in any group activity class. The common features prevent discriminating between different group activities.
% 
% For example, between 2p-succ and 3p-succ, the post-shot behavior (e.g., waiting for the restart of a play) is similar.
% 
% In such cases, 1-NN is sensitive to the common features, whereas KNN with a larger $K$ can avoid the negative impact with the ensemble of $K$ neighborhoods, which explains the larger performance gains on NBA when increasing $K$.

\begin{table}[t]
    \centering
    \caption{Effect of temporal resolutions on NBA. MCA/MPCA are shown.
    %We obtained the scores using the authors' official code for~[\textcolor{cvprblue}{21, 26}].
    For fair comparison under equivalent conditions, we select~[\textcolor{cvprblue}{21, 26}] whose authors provide publicly available code.
    }
    % \vspace*{-3mm}
    \scalebox{0.85}{
    % {\renewcommand{\arraystretch}{0.95}
    \begin{tabular}{l|cccc}
    \hline
    Method & 3 frames                 & 6 frames                 & 12 frames                & 18 frames                \\ \hline
    DFWSGAR~[\textcolor{cvprblue}{21}]                 & 65.3/60.2          & 71.8/65.4          & \textbf{74.4/69.4} & 73.9/68.6          \\
    MP-GCN~[\textcolor{cvprblue}{26}]                  & 49.4/45.8          & 63.7/59.3          & 70.8/66.6          & \textbf{74.4/70.8} \\
    Ours                    & \textbf{67.4/61.4} & \textbf{73.1/67.7} & 73.0/68.2          & 73.2/68.4          \\ \hline
    \end{tabular}
    }
    % }
    \vspace*{-2mm}
    \label{tab:NBA_wegar_frame}
\end{table}

\subsubsection{Fine-tuning for Supervised Recognition}
\label{subsubsection:finetune_sgar}
%
% While the results of the supervised recognition on VBD are shown in Table~\textcolor{cvprblue}{5} of the main paper, we further show the results on NBA in Table~\ref{tab:comp_wsgar_nba}.
While the results of the supervised recognition on terms of MCA on both VBD and NBA are shown in Table~\textcolor{cvprblue}{5} of the main paper, we further show the results of Merged MCA on VBD and MPCA on NBA in Table~\ref{tab:comp_wsgar_nba}. Merged MCA~[\textcolor{cvprblue}{56}] is an evaluation metric in which the pass and set classes are merged into one class.
Although our method achieves state-of-the-art performance on VBD, it is not the best on NBA.
% 
% However, our method remains competitive with several baselines such as SAM~[\textcolor{cvprblue}{56}], Dual-AI~[\textcolor{cvprblue}{13}], and LiGAR~[\textcolor{cvprblue}{4}], despite using a relatively simple architecture and a generalized pretrained feature extractor.
% 
One possible reason for the low performance on the NBA is the dataset's characteristics. 
Compared to VBD, each sequence in NBA is longer, and a group activity label (e.g., 2p-succ, 3p-succ) often spans a long temporal interval. 
In our method, we process the sequence using a temporal Transformer and then apply max pooling over time to obtain $\bm{G}$. 
As a result, features of all frames are compressed into $\bm{G}$, which prevents us from discriminating between similar group activities.
For example, 2p-succ and 3p-succ are similar because of players’ behaviors after a successful shot.
This may partly explain why our generic architecture is less effective than NBA-specific designs such as SOGAR~[\textcolor{cvprblue}{6}], Flaming-Net~[\textcolor{cvprblue}{29}], and MP-GCN~[\textcolor{cvprblue}{26}], which explicitly incorporate richer temporal modeling specialized for NBA.
These results suggest that our GAFs can be improved by more flexible temporal aggregation modules as used in these methods.
We leave exploring such NBA-specific temporal modules for future work, while preserving the simplicity of our generic GAF learning network.
%
% [\R1] \underline{\textbf{Temporal interval analysis.}}
% To support our interpretation of the performance drop on NBA (i.e., unsuitable pooling over a long time) in Sec.~\textcolor{cvprblue}{7.2.2} of the Supp, Table~\ref{tab:NBA_wegar_frame} shows results for input videos with different frame counts.

This interpretation is further supported by Table~\ref{tab:NBA_wegar_frame}, which shows results for input videos with different frame counts.
While keeping the elapsed time per video unchanged, we vary the number of frames by changing the temporal resolution.
Our method shows an advantage in low-frame settings, where the negative impact of pooling is limited because fewer frames are pooled.

\subsection{Visualizations}
\label{subsection:sup_vis}

In addition to the visualization of GAFs shown in Fig.~\textcolor{cvprblue}{10} of the main paper, we also visualize actual images corresponding to GAFs in Fig.~\ref{fig:vis_fine}.

First, we can see that contextually similar group activities are located in similar regions in Fig.~\ref{fig:vis_fine}. For example, R-set data points (i.e., (c) and (d)) and R-pass data points (i.e., (e) and (f)) are close in the learned GAF space. This is because both group activities include the player interacting with the ball for the future spike. Moreover, we can also see that R-set data points (i.e., (c) and (d)) and R-spike data points (i.e., (a) and (b)) are close in the learned GAF space. These results also verify that our method can learn contextual similarity between R-set and R-spike activities.
Based on the discussion above, we can summarize that our GAF learning can capture class-level transitions (e.g., offense progression from Pass to Set to Spike).

Furthermore, Fig.~\ref{fig:vis_fine} shows that our method can learn fine-grained similarities among various group activities beyond the class-level similarities.
% 
% In Fig.~\ref{fig:vis_fine_R_spike}, the orange data points (i.e., R-spike) split into two clusters. These clusters reflect whether the spiker hits the ball from different areas on the court.
First, it can be seen that the orange data points (i.e., R-spike) split into two clusters (a) and (b). While the spiker hits the ball at the top of the image in (a), the spiker hits the ball at the bottom of the image. These differences in spiker position in (a) and (b) demonstrate that our method learns fine-grained information of players who are important for representing group activity (e.g., spikers in these examples).

% Figure~\ref{fig:vis_fine_R_set_pass} visualizes how the green data points (i.e., R-pass) and the blue data points (i.e., R-set) are distributed in the feature space. The R-pass examples, far from the blue cluster in (d), correspond to a scene where the receiver passes the ball far from the net, that is, in a formation still distant from an attacking formation. 
% 
% In contrast, the R-pass examples, located near the blue cluster in (c), show a scene where the ball is received near the center of the court. 

In (c) and (d), blue data points (i.e., R-set) are split into two clusters.
While the players are tossing the ball near the net in (c), the players are tossing the ball far from the net in (d). These examples also validate that our GAF learning preserves the position information of players, as in (a) and (b).
In (e) and (f), green data points (i.e., R-pass) are split into two clusters. In (e), the players are receiving the ball near the center of the court. In (f), we can observe that the players are receiving the ball at a distance from the net. 

Interestingly, we can also see that (b) and (c) are close in the learned GAF space. The closeness comes from the position similarity of players interacting with the ball. As with these results, (d) and (e) are close in the learned GAF space. This is because both examples contain players interacting with the ball at a similar position. These results demonstrate that our method can learn fine-grained similarities among group activities beyond class-level similarities.

% The R-set examples close to the green cluster in (b) correspond to a toss made near the center of the court. Finally, the R-set examples, far from the green cluster and close to the orange data points (i.e., R-spike) in (a), show a scene where the toss is made near the net, and the team is already in an attacking formation that is likely to lead to a spike.

% This distribution of samples can be interpreted as follows. R-pass and R-set are arranged in the GAF space along a continuous axis reflecting the distance from the net and the progression toward an attacking formation. 
% 
% Passes far from the net (d) appear at one end of this axis, whereas passes near the court center (c) lie closer to the R-set cluster. Moreover, R-set samples gradually move toward the R-spike cluster from (b) to (a) as the toss position shifts from the court center to near the net. These observations indicate that our GAFs capture not only class-level distinctions but also fine-grained transitions in offensive formations, such as the progression from receiving to setting to spiking.


\end{document}